\documentclass[twoside,leqno,twocolumn]{article}
\usepackage{ltexpprt}

\usepackage{graphicx}
\usepackage{float}
\restylefloat{figure}
\usepackage{subfigure}
\usepackage{color}
\usepackage{amsmath}
\usepackage{amsfonts}
\usepackage{amssymb}
\usepackage{booktabs}
\usepackage{multirow}
\usepackage{rotating}
\usepackage{url}

\begin{document}

\title{\Large A Full Probabilistic Model for Yes/No Type Crowdsourcing\\in Multi-Class Classification}
\author{Belen Saldias-Fuentes\thanks{Computer Science Department,  Pontificia Universidad Cat\'olica de Chile, Santiago, Chile.} \thanks{MIT Media Lab,  Massachusetts Institute of Technology, Cambridge, MA, USA (present affiliation, belen@mit.edu).}\\
\and
Pavlos Protopapas\thanks{Institute for Applied Computational Science, Harvard University, Cambridge, MA, USA.}\\
\and
Karim Pichara B.\footnotemark[1]  \footnotemark[3]
}
\date{}

\maketitle


\fancyfoot[R]{\scriptsize{Copyright \textcopyright\ 2019 by SIAM\\
Published under the terms of the Creative Commons 4.0 license.}}





\begin{abstract} \small\baselineskip=9pt 
\fontsize{8.5pt}{9pt}\selectfont
Crowdsourcing has become widely used in supervised scenarios where training sets are scarce and difficult to obtain. Most crowdsourcing models in the literature assume labelers can provide answers to full questions. In classification contexts, full questions require a labeler to discern among all possible classes. Unfortunately, discernment is not always easy in realistic scenarios. Labelers may not be experts in differentiating all classes. In this work, we provide a full probabilistic model for a shorter type of queries. Our shorter queries only require ``yes'' or ``no'' responses. Our model estimates a joint posterior distribution of matrices related to labelers' confusions and the posterior probability of the class of every object. We developed an approximate inference approach, using Monte Carlo Sampling and Black Box Variational Inference, which provides the derivation of the necessary gradients. We built two realistic crowdsourcing scenarios to test our model. The first scenario queries for irregular astronomical time-series. The second scenario relies on the image classification of animals. We achieved results that are comparable with those of full query crowdsourcing. Furthermore, we show that modeling labelers' failures plays an important role in estimating true classes. Finally, we provide the community with two real datasets obtained from our crowdsourcing experiments. All our code is publicly available\footnote{\url{https://github.com/bcsaldias/yes-no-crowdsourcing}}.
\end{abstract}

\section{Introduction.}
\label{sec:Introduction}
Labeled data is the very first requirement for training classifiers. Moreover, the availability of data has stimulated great breakthroughs in AI. For example, convolutional neural networks (CNNs) were first proposed by \cite{lecun1989backpropagation}, but only when ImageNet \cite{deng2009imagenet} achieved a corpus of 1.5 million labeled images could Google's GoogLeNet \cite{krizhevsky2012imagenet} perform object classification almost as well as humans by using CNNs. This encouraged us to create new mechanisms for producing labels. Nevertheless, labeling means getting ground truths, which are often difficult, expensive, or impossible to obtain.

To increase the amount of labeled data, we can use crowdsourcing \cite{dawid1979maximum, raykar2010learning, simpson2013dynamic, vaughan2018making} to gather a large amount of labels. A major challenge is to combine unreliable crowd information: this is not entirely accurate, but cheaper \cite{yan2012modeling}. A typical case is to take the majority of votes for each object. For this to work, we must assume everyone has equal knowledge about the topic, which is in many cases a wrong assumption. In addition, we can use active learning (AL) \cite{zhang2015active, yan2011active}, a semi-supervised scenario in which a learning model iteratively selects the best instances (for example, those that most confuse the model) to be tagged by an expert. We can also mix these strategies \cite{zhang2015active, liu2012truelabel+, yan2012modeling} to select candidates by considering labelers' expertise. Nevertheless, here we propose a model to make the labeling task even easier.

\begin{figure}[!h]
\centering    
\subfigure[Query types]{
    \label{fig:strategies}
	\includegraphics[width=0.47\linewidth]{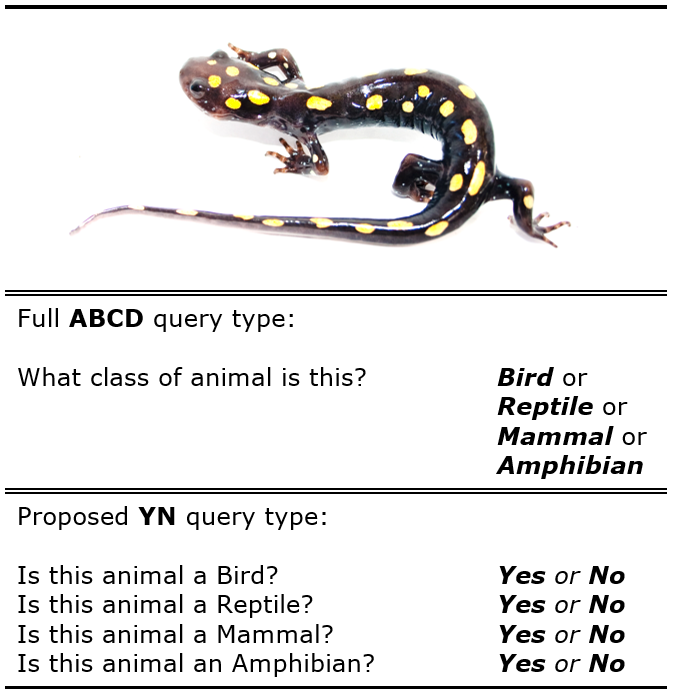} 
}
\subfigure[Questions for a crowd]{
   \label{fig:different-labelers}
  \includegraphics[width=0.451\linewidth]{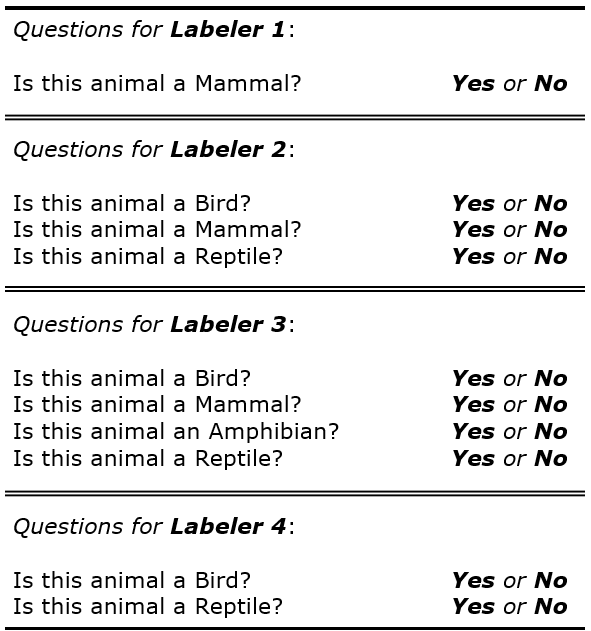}
}
    \caption[Different query scenarios]{Different query scenarios. Figure \ref{fig:strategies} shows the spotted salamander, an amphibian. Figure \ref{fig:different-labelers} shows a possible scenario with four labelers, four classes, and ``yes'' or ``no'' questions for the animal in \ref{fig:strategies}.}
   \label{fig:query-type}
\end{figure}

Instead of selecting the best instances as candidates for training the model, we propose a novel approach to query type (see figure \ref{fig:query-type}). Typically in a four-class scenario, a labeler is asked the class of an object with possible responses ``A'' or ``B'' or ``C'' or ``D''. We refer to that type of full question as an ABCD question. Our model generates low-cost queries in which each response gives partial information. This method iteratively selects, per labeler, a random object along with a class' label, then asks if that object belongs to that class: ``yes'' or ``no'' (proposed YN question).

The proposed method has many advantages over traditional approaches. First, the YN model focuses on the importance of learning an estimation of how labelers fail. Our strategy probabilistically learns initial parameters from the data for the labeling stage. Second, the labelers do not need to know all the classes. Third, it captures partial information with fewer errors because the labelers do not need to know the ground truth to accurately respond to some YN questions. Finally, the method is independent of the kind of data, given that we only need to include labelers' votes, without worrying about representation of the objects to be classified.

This work makes the following main contributions:
\begin{enumerate}
\item \textit{Crowdsourcing query type}: We propose a new crowdsourcing framework to obtain labeled data focused on the query type. This method costs less than other models because it reconstructs ground truth labels by only using partial information. We show that the aggregation of partial information allows the YN model to ask fewer questions than others, while achieving similar accuracy.
\item \textit{New data released}: We developed two real-world experiments with humans and published the data.
\end{enumerate}

The rest of this paper is organized as follows: Section \ref{sec:related-work} presents some related work. In section \ref{sec:the-model} we explain the proposed model, and in section \ref{sec:Inference-Scheme} we show how we solved it. Then, section \ref{sec:implementation} describes our implementations of the model. Section \ref{sec:data} describes the datasets for comparison. Then, section \ref{sec:results} shows experiments and analysis. Finally, in section \ref{sec:conclusions} we discuss and conclude with the main results of our work.

\section{Related Work}
\label{sec:related-work}
\subsection{Creating Training Sets}
To acquire labels, we can manually label as many objects as possible. Furthermore, others have used crowdsourcing or/and active learning \cite{zhang2015active, liu2012truelabel+, yan2012modeling}. From another point of view, \cite{RatnerCreatingLarge} proposes using \textit{data programming}, in which labelers give functions that return the asked labels. Another option to create labels is \textit{co-training} \cite{blum1998combining}, in which data is labeled from two independent views. Closer to our approach is \textit{boosting} \cite{schapire2012boosting}, which combines several ``weak'' classifiers to create a ``strong'' one. We considered the weaknesses by modeling the labelers' (many views) errors to infer the true labels probabilistically.

\subsection{Crowdourcing Scenarios}
\label{subsec:crowds}
Several efforts have been made on estimating labelers' expertise \cite{yan2010modeling,yan2011active,liu2012truelabel+,simpson2013dynamic} and maximizing labelers' accuracy by giving them the right incentives \cite{shah2015double}. Some researchers have proposed new query types on active learning scenarios \cite{rashidi2011ask, huang2015multi}. Additionally, there are strategies to optimize the trade-off between redundancy and reliability in multi-class scenarios \cite{karger2013efficient}. The closest research to the YN query type \cite{qi2008two} involved assuming that each instance could belong to more than one class. However, these works did not involve a crowdsourcing context to improve the scenario. They mostly maintained a perfect oracle assumption.

Until now, no research has been presented to integrate query type, partial information asked to labelers, and the power of crowd. We propose a mechanism that outperforms other methods and handles many difficulties, as we outlined in section \ref{sec:Introduction} and through this paper.

\subsection{Variational Inference Approaches}
\label{subsec:vi}
Several inference schemes have been used to solve the YN model. Following a probabilistic perspective, EM or MAP algorithms make the YN model very likely to converge to a local optimum \cite{raykar2010learning, yan2012modeling}. This can be handled using the Gibbs sampler \cite{geman1984stochastic, liu2012truelabel+}. Previous research on labeling has always involved methods for full questions.

We used the No-U-Turn Hamiltonian sampler (NUTS) \cite{hoffman2014no} to converge more quickly than the random walk that MCMC \cite{hastings1970monte,gelfand1990sampling} uses. Additionally, we tested Black Box variational inference (BBVI) \cite{ranganath2014black}  because it tends to be faster than NUTS \cite{simpson2013dynamic}. BBVI is inexpensive and easy to implement because it only requires estimating the ELBO gradient. 


\section{The Model}
\label{sec:the-model}
Consider a dataset with $\mathcal{N}$ objects; each object $\mathcal{X}_i$ has only one true class $z_i$, among $\mathcal{K}$ possible classes, where $i \in \{1,...,\mathcal{N}\}$ and $\mathbf{Z} = \{z_1,..., z_i,...,z_\mathcal{N}\}$. Each labeler $\mathcal{L}_j$ is then presented with a series of binary ``yes'' or ``no'' (YN) questions, where $j \in \{1,...,\mathcal{J}\}$.

Formally, we define a YN question $k_i^j$ as the question asked to labeler $\mathcal{L}_j$ about whether $\mathcal{X}_i$ belongs (``yes'' or ``no'') to the class $M_k$, $k \in \{1,...,\mathcal{K}\}$. We define $\mathcal{K}_i^j$ as the set of $k_i^j$ queries asked to labeler $\mathcal{L}_j$ for the object $\mathcal{X}_i$. Let $r_{ik}^j$ be the response (or vote) assigned by $\mathcal{L}_j$ to the question $k_i^j$, and $\mathbf{R}$ the set of all responses $r_{ik}^j$. Note that a labeler is not asked twice for the same class for the same object.

\begin{figure}[!h]
    \centering
    \includegraphics[width=.75\linewidth]{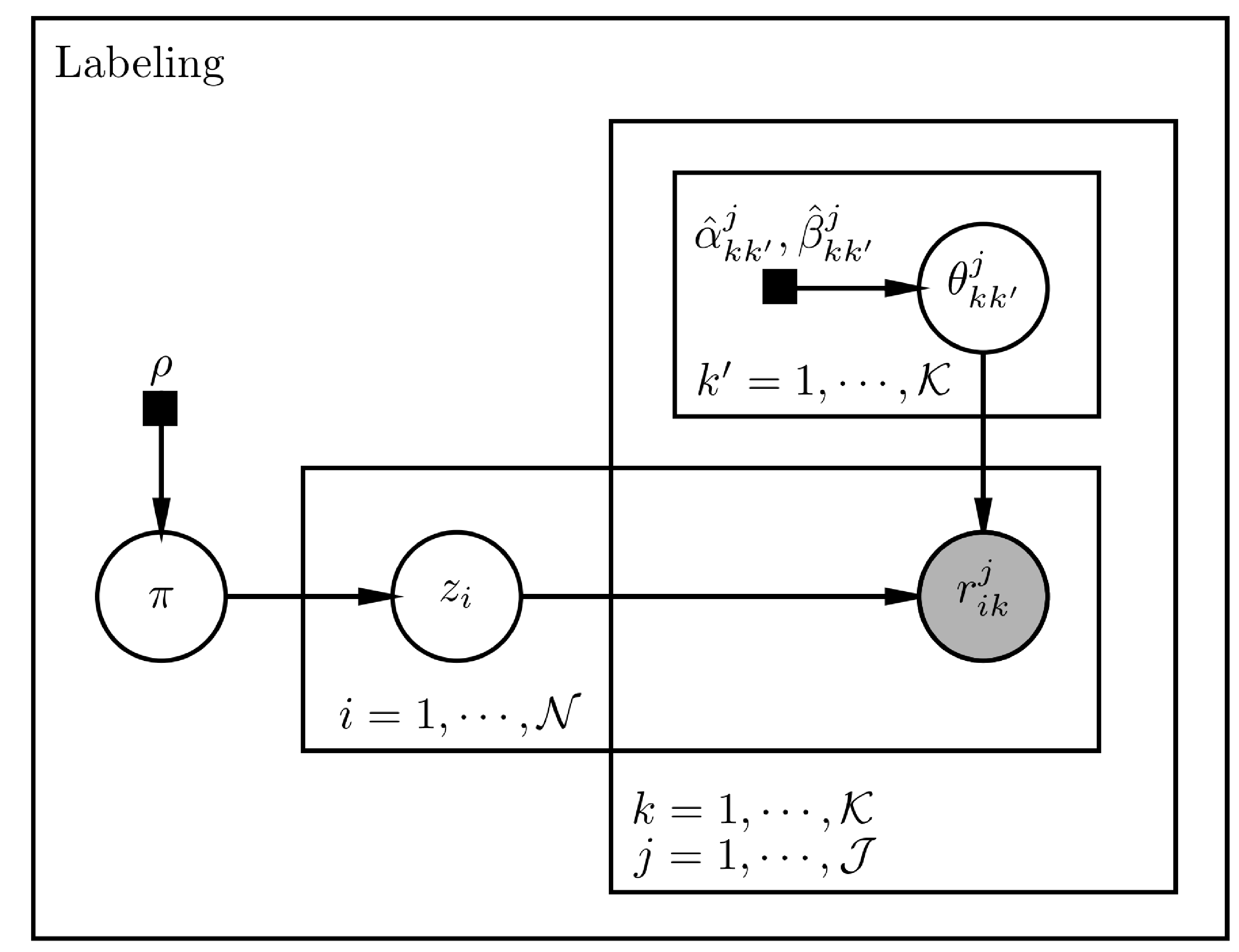}
    \caption{Proposed PGM. In this plate notation, random variables are clear circles; observed variables are shaded in gray. Estimated prior hyperparameters are represented by squares.}
    \label{fig:the-model}
\end{figure}

We propose a probabilistic graphical model \cite{koller2009probabilistic, wainwright2008graphical} (shown in figure \ref{fig:the-model}) to infer the true labels $\mathbf{Z}$. The \texttt{Labeling} area represents the joint distribution of  $\mathbf{Z}$ and the other variables involved in their prediction.

\subsection{Responses}
\label{sec:responses}
For object $\mathcal{X}_i$, labeler $\mathcal{L}_j$, and question $k_i^j$, it is convenient to encode the response as a two dimensional vector: $r_{ik}^{j}$, where $[0,1] \leftarrow$ [YES, NO]. Figure \ref{fig:voter-table} shows an example of votes for object $\mathcal{X}_i$ given by labeler $\mathcal{L}_j$. Note that $r_{ik}^{j} = [0, 0]$ means that question $k_i^j$ was not asked. 

\begin{figure}[h]
\centering
\small\addtolength{\tabcolsep}{-2pt}
\centering
{\renewcommand{\arraystretch}{1.05}		
\centering
\begin{tabular}{c c c}    \toprule
\emph{Question for} &   \emph{Yes} &  \emph{No}  \\\midrule
 Class $\mathcal{M}_1$                & 1            & 0           \\
Class $\mathcal{M}_2$                 & 0            & 0           \\
\vdots                  & \vdots       &    \vdots   \\
Class $\mathcal{M}_k$               & 0            & 1           \\
\vdots                  & \vdots       &    \vdots   \\
Class $\mathcal{M}_\mathcal{K}$     & 0            & 1  \\\bottomrule
 \hline
\end{tabular}}

\caption{Responses/votes $r_{i}^{j}$.}
\label{fig:voter-table}
\end{figure}
\subsection{Credibility Matrices}

Common approaches involve the use of the confusion matrix of each labeler to represent their errors, due to the nature of the full question. We represented the YN error per labeler as a \textit{credibility matrix}. We needed to find the probability per labeler of giving the right answer when the class asked is $M_{k'}$, and the true class is $M_k$. Figure \ref{fig:credibility-matrix-table} shows the credibility matrix of a specific labeler, where $\theta_{kk'}^{j}$ is the probability of labeler $\mathcal{L}_j$ saying ``yes'' to question ${k'}_{i}^{j}$ when $z_i = k$. We assumed that the labelers were not random voters so that we could find patterns in their behaviors.

Our main goal was to find the most likely class for each object, given the votes and \textit{credibility matrices} $\Theta$. A side goal was to estimate $\Theta$. In particular, we considered conjugate priors. Given that each ``yes'' or ``no'' $r_{kk'}^j$ response can be modeled as a $\mathrm{Bernoulli}$ distribution, the prior for $\theta_{kk'}^{j}$ distributes $\mathrm{Beta}(\hat{\alpha}_{kk'}^{j},\hat{\beta}_{kk'}^{j})$, where $\hat{\alpha}_{kk'}^{j}$ and $\hat{\beta}_{kk'}^{j}$ are the estimated prior initial parameters from the first stage. Finally, the likelihood is:
$$r_{kk'}^j \sim \mathrm{Bernoulli}(\theta_{kk'}^{j})$$

Modeling the prior of $\theta_{kk'}^{j}$ as a $\mathrm{Beta}$ distribution that lives in a 0 to 1 space allowed us to model the probability of a response. It is also a conjugate distribution for the Bernoulli likelihood and can model any expertise due to its flexibility.
\begin{figure}[h]
\small\addtolength{\tabcolsep}{-2pt}
\centering

{\renewcommand{\arraystretch}{1.35}	
\centering
\begin{tabular}{c c c c c c c} \toprule
\multicolumn{1}{c}{}  & \multicolumn{4}{c}{}\emph{Question for $\mathcal{M}_{k'}$} \\[1.5pt]

\multicolumn{1}{c}{}                                          & \multicolumn{1}{c}{$\theta_{1,1}$}           & \multicolumn{1}{c}{$\theta_{1,2}$}           & \multicolumn{1}{c}{\ldots} & \multicolumn{1}{c}{$\theta_{1,k'}$}           & \multicolumn{1}{c}{\ldots} & \multicolumn{1}{c}{$\theta_{1,\mathcal{K}}$}           \\  

\multicolumn{1}{c}{}                                          & \multicolumn{1}{c}{$\theta_{2,1}$}           & \multicolumn{1}{c}{$\theta_{2,2}$}           & \multicolumn{1}{c}{\ldots} & \multicolumn{1}{c}{$\theta_{2,k'}$}           & \multicolumn{1}{c}{\ldots} & \multicolumn{1}{c}{$\theta_{2,\mathcal{K}}$}           \\  

\multicolumn{1}{c}{}                                          & \multicolumn{1}{c}{\vdots}                   & \multicolumn{1}{c}{\vdots}                   & \multicolumn{1}{c}{$\ddots$} & \multicolumn{1}{c}{\vdots}                    & \multicolumn{1}{c}{$\ddots$} & \multicolumn{1}{c}{\vdots}                             \\  

\multicolumn{1}{c}{}                                          & \multicolumn{1}{c}{$\theta_{k,1}$}           & \multicolumn{1}{c}{$\theta_{k,2}$}           & \multicolumn{1}{c}{\ldots} & \multicolumn{1}{c}{$\theta_{k,k'}$}           & \multicolumn{1}{c}{\ldots} & \multicolumn{1}{c}{$\theta_{k,\mathcal{K}}$}           \\  

\multicolumn{1}{c}{}                                          & \multicolumn{1}{c}{\vdots}                   & \multicolumn{1}{c}{\vdots}                   & \multicolumn{1}{c}{$\ddots$} & \multicolumn{1}{c}{\vdots}                    & \multicolumn{1}{c}{$\ddots$} & \multicolumn{1}{c}{\vdots}                             \\  
\multicolumn{1}{c}{\multirow{-6}{*}{
\begin{turn}{90}
\emph{True Class $\mathcal{M}_k$}
\end{turn}}} & \multicolumn{1}{c}{$\theta_{\mathcal{K},1}$} & \multicolumn{1}{c}{$\theta_{\mathcal{K},2}$} & \multicolumn{1}{c}{\ldots} & \multicolumn{1}{c}{$\theta_{\mathcal{K},k'}$} & \multicolumn{1}{c}{\ldots} & \multicolumn{1}{c}{$\theta_{\mathcal{K},\mathcal{K}}$} \\ \bottomrule 
\end{tabular}
}

\caption{Credibility matrix. Note that the rows are not required to sum 1.}
\label{fig:credibility-matrix-table}
\end{figure}

\subsection{Joint Distribution}
\label{subsec:joint-distribution}
Each YN vote $r_{ik}^{j}$ depends on the real, but unknown, label  $z_i$. Furthermore, the vote also depends on the credibility $\theta_{z_ik}^{j}$ of labeler $\mathcal{L}_j$. The conditioning to $z_i$ allows the labeler to be more accurate in subsets of classes. The dependency on $\Theta^j$ allowed us to model the labeler's biases and errors for all classes. These dependencies are represented by the conditional distribution $\mathrm{P}(r_{ik}^{j} | z_i, \theta_{z_ik}^j)$ \cite{liu2012truelabel+}.

From prior information, we could estimate the initial class proportions $\rho$ and define a global Dirichlet variable $\pi$ in charge of this unknown distribution of vector $\mathbf{Z}$. Finally, this gave:
$$\pi \sim \mathrm{Dirichlet}(\rho)$$
$$z_i \sim \mathrm{Categorical} (\pi)  $$

\subsubsection*{Likelihood} We started from a single labeler, one object, and one question. For labeler $\mathcal{L}_j$ and question $k_{j}^{j}$, the likelihood is found in (\ref{eq:like0}), where we encoded the response as a two-dimensional vector: $r_{ik}^{j}$, where $[0,1] \leftarrow$ [YES, NO].  For all responses $\mathbf{R}$, all labelers $\mathcal{L}$, and all data $\mathcal{N}$, the likelihood is found in (\ref{eq:like1}). 

\begin{equation}
\footnotesize
\mathrm{P}(r_{ik}^{j} | \theta_{z_ik}^j, z_i) = \overbrace{\overbrace{(\theta_{z_ik}^j)^{r_{ik}^j[0]}}^{r_{ik}^{j}[0]=1, r_{ik}^{j}[1]=0 }}^{YES} \times  \overbrace{\overbrace{ (1-\theta_{z_ik}^j)^{r_{ik}^j[1]}}^{r_{ik}^{j}[0]=0,  r_{ik}^{j}[1]=1}}^{NO}
\label{eq:like0}
\end{equation}

\begin{equation}
\footnotesize
\mathrm{P}(\mathbf{R} | \Theta, \mathbf{Z}) \propto \prod_{i=1}^{\mathcal{N}} \,\, \prod_{j=1}^{\mathcal{J}} \,\, \prod_{k \in \mathcal{K}_i^{j}}
    \left\{  {(\theta_{z_ik}^{\,j}})^{r_{ik}^{j}[0]}\,(1-\theta_{z_ik}^{\,j})^{r_{ik}^{j}[1]}
    \right\}
    \label{eq:like1}
\end{equation}

\section{Inference Schema}
\label{sec:Inference-Scheme}

We separated the inference into two intuitive stages: first, to estimate the labelers' reliability by asking them for known objects $\hat{\mathcal{N}}$ (Training Set), and second to ask them for unknown objects labels. We could unify these stages in a single inference model with an identical result. In the scenario where $\hat{\mathbf{Z}}$ are observed values, the model estimates beforehand $\hat{\Theta}$ and converges faster (see section \ref{sec:results}). The likelihood for all responses $\hat{\mathbf{R}}$, all labelers $\mathcal{L}$, and all data $\hat{\mathcal{N}}$ is found in (\ref{eq:fulllike}).

\begin{equation}
\footnotesize
\mathrm{P}(\hat{\mathbf{R}}, \hat{\mathbf{Z}}\,| \hat{\Theta}) \propto \prod_{i=1}^{\hat{\mathcal{N}}} \,\, \prod_{j=1}^{\mathcal{J}} \,\, \prod_{k \in \hat{\mathcal{K}}_i^{j}}
    \left\{  {(\hat{\theta}_{\hat{z}_ik}^{\,j}})^{\hat{r}_{\hat{i}k}^{j}[0]}\,(1-\hat{\theta}_{\hat{z}_ik}^{\,j})^{\hat{r}_{\hat{i}k}^{j}[1]}
    \right\}
    \label{eq:fulllike}
\end{equation}

The prior distribution of each $\hat{\theta}$ was chosen to be uninformative, but flexible enough to represent labelers with both high and low expertise. We selected $\hat{\theta}_{kk'}^{j} \sim \mathrm{Beta}(\alpha,\beta)$ with an expected value equivalent to $0.5$ (see section \ref{sec:results}). As stated before, this inference scheme works in two stages (that can also be done analytically):
\begin{enumerate}
\item \texttt{Credibility} stage: estimating $\hat{\Theta}$. Because we assumed the labelers would behave similarly in the \texttt{Labeling} stage, as they do here, we obtained the $\hat{\alpha}_{kk'}^{j}$ and $\hat{\beta}_{kk'}^{j}$ parameters from each $\hat{\theta}_{kk'}^{j}$.
\item \texttt{Labeling} stage: predicting $\textbf{Z}$ and $\Theta$ via posterior inference.
\end{enumerate}


\section{Implementation}
\label{sec:implementation}
Due to the convergence time of NUTS, we also used BBVI \cite{ranganath2014black}, both in Python3.5. Each one works as follows: First, it estimates the latent variables $\hat{\Theta}$. Second, it estimates $\Theta$, $\mathbf{Z}$, and $\pi$. All the experiments presented in section \ref{sec:results} used NUTS \cite{pymc3}, except when indicated otherwise. BBVI approximately tries to find a probability distribution that is closest (in KL divergence) to the true posterior distribution. The supplementary material provides the derivation of the needed gradients to solve the model, which can be easily extended to any model with similar variable types (based on \cite{chaney2015guide}). 
\section{Data}
\label{sec:data}
We used simulated and real-world datasets. First, we simulated data to understand the YN model's behavior. Then, we trained classifiers with real-world data to produce responses and evaluate the YN model performance. Finally, we tested the model in two human scenarios. These three sources of labels are described in the following subsections.

\subsection{Synthetic Votes for Synthetic Data.} 
\label{data:synthetic}
To simulate labelers and their votes, we proceeded as follows: First, we created labels ($\mathbf{\hat{Z}}$ and $\mathbf{Z}$). Then, for each labeler, we sampled a credibility matrix. Each row was simulated using a $\mathrm{Beta(0.5, 0.5)}$ distribution. Labelers have high expertise in at most half of the classes; expertises were sampled from a $\mathrm{Beta(20, 1)}$ distribution (because its expected value is close to 1). Finally, we simulated the votes using the labelers and true labels. When the labeler $\mathcal{L}_j$ was presented with object $\mathcal{X}_i$ of class $z_i$ and is asked $k_{i}^{j}$, we consulted its credibility matrix to obtain the response for $k_{i}^j$. We took $r_{ik}^j$ by flipping a coin with the probability given by $\theta_{z_ik}^{\,j}$.

\subsection{Synthetic Votes for Real-World Data.}
\label{data:macho}
We used a subset of MACHO data \cite{cook1995variable} (250 objects). We trained six different classifiers as labelers, each with a different training set but equally sized (2 Random Forest classifiers, 2 Logistic Regressions, and 2 Support Vector Machines). We proceeded as follows: First, we split the data into three different sets; one to train classifiers, another to infer $\hat{\Theta}$, and the last to test the model. Each labeler was composed of a pool of $\mathcal{K}$ one-vs-all classifiers. When a labeler was asked for $k_{i}^j$, we consulted its one-vs-all binary classifier for the class $\mathcal{M}_k$ to get the probability of the object belonging to the class $M_k$. Then, we flipped a coin with that probability to obtain $\mathbf{\hat{R}}$ and $\mathbf{R}$.
\begin{itemize}
  \item[] MACHO data: Irregularly-sampled time series. Several works aim to classify astronomical irregular time series \cite{pichara:2016}. Table \ref{tab:all-data} shows the data distribution that we used. 
\end{itemize}

\subsection{Real Votes for Real-World Data.} 
\label{data:websites}
Two websites were set up to acquire data from human crowds. Each of them presented a contest to people related to a specific dataset domain (see table \ref{tab:all-data}):

\begin{enumerate}
\item Astronomical irregular time series: We aim to classify irregular time series of the Catalina Surveys \cite{drake2009first}. The labelers, 8 in total, were astronomers and engineers familiar with the field. From the human experiments, we proved that our model can assist astronomers' work. 

\item Animal classes: The objective of classifying animals\footnote{The full dataset is available at: https://a-z-animals.com/animals/pictures/. We filtered the number of mammals to do not have an extremely unbalanced dataset. The class fish was removed to work with only four classes and to increase the difficulty.} was to compare the model in different fields. The labelers selected were 11 university students.
\end{enumerate}

Each dataset contains 4 classes and 318 unknown objects, for about 15 people. Each user was presented with 1 to 4 random YN questions per instance. Also, the sets have (i) 40 and (ii) 41 known objects, respectively. For those known objects and 80 of the 318 unknown ones, the users were asked the ABCD question as well. The following results are based only on those labelers who finished at least 70\% of the questions.

\begin {table}[h]
\caption {Instances per class for each real-world dataset.} \label{tab:all-data} 
\centering
{\renewcommand{\arraystretch}{1.1}		
\centering
\small\addtolength{\tabcolsep}{-3pt}
\begin{tabular}{ccccccc}    \toprule
\multicolumn{2}{c}{\emph{MACHO}} &   \multicolumn{2}{c}{\emph{The Catalina Surveys}} &  \multicolumn{2}{c}{\emph{Animals}}  \\\midrule
EB      &   104      & \multicolumn{1}{|c}{CEP}      & 119      & \multicolumn{1}{|c}{Mammal}     & 232      \\
BE      &   57       & \multicolumn{1}{|c}{RRLYR}   & 99         & \multicolumn{1}{|c}{Bird}    & 73       \\
LPB     &   49       & \multicolumn{1}{|c}{EB}      & 80     & \multicolumn{1}{|c}{Amphibian}    & 31       \\
CEP     &	40	   & \multicolumn{1}{|c}{LPV}      & 60      & \multicolumn{1}{|c}{Reptile}  & 23 \\\bottomrule
 \hline
\end{tabular}}
\end{table}
\vspace{-0.25em}

\section{Results}
\label{sec:results}
The experiments are divided into eleven parts: Two full experiments with synthetic data (\ref{result:1} and \ref{result:2}); four using classifiers on MACHO data (\ref{result:3}, \ref{result:4}, \ref{result:5}, and \ref{result:6}); finally, we set up the websites to get real crowds' results, which we present in five experiments (\ref{result:7}, \ref{result:8}, \ref{result:9}, \ref{result:10}, and \ref{result:11}). We used NUTS for all experiments, except for the benchmark against BBVI presented in experiment \ref{result:7}. We always used ten sampling chains and \textit{burned} the first 1500 samples.

\subsection{Convergence Simulations - Synthetic Data.}
\label{result:1}
We created votes, as explained in subsection \ref{data:synthetic}. For synthetic and classifiers' votes, we used six labelers and four classes. We asked each labeler between 1 and 4 questions (Random(1,4)) for about 250 objects. Between 25 and 40 objects were used to approximate $\hat{\Theta^j}$; the rest were used for testing.

For all experiments we performed, the classification accuracy scores became completely stable after 3000 iterations. Similar results for convergence were obtained from both classifiers' scenarios and the two set-up contests with real-world data. The convergence of each variable ($\hat{\Theta}$, $\pi$, $\textbf{Z}$, and $\Theta$) was diagnosed based on the Gelman-Rubin statistic \cite{gelman1992inference}. They all converged.


\subsection{Modeling the Crowd Expertise - Synthetic Data.}
\label{result:2}
\begin{figure}[!ht]
    \centering
    \includegraphics[width=\linewidth]{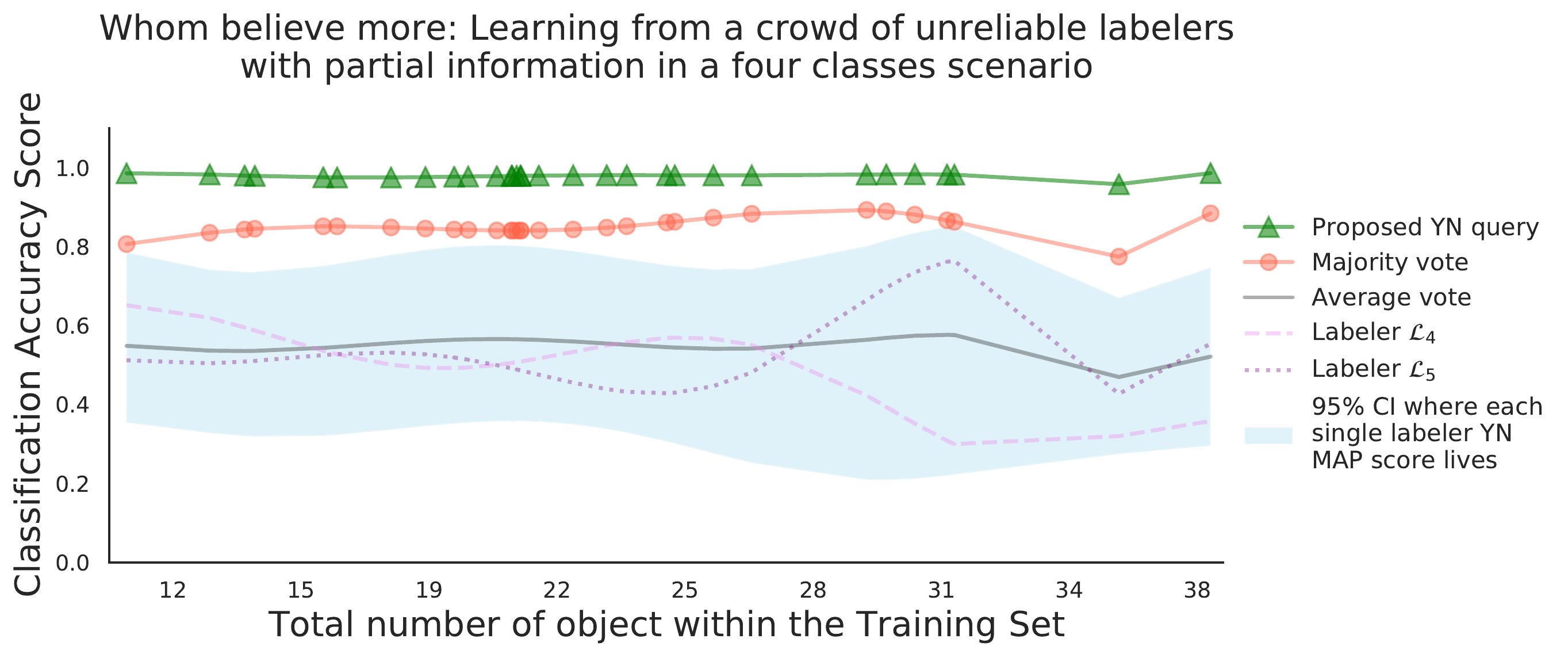}
    \caption[Crowd expertise on synthetic data]{Crowd expertise on synthetic data. Note that each labeler score lies in a range of lower accuracy classification score than the YN and majority methods.}
    \label{fig:good-bad-experts}
\end{figure}

To prove that our model can effectively differentiate between accurate and inaccurate labelers, we compared it with the baselines used in \cite{yan2010modeling}. Here, we worked with 7 synthetic labelers with higher expertise for at most two of four classes (as explained in section \ref{sec:data}). Figure \ref{fig:good-bad-experts} shows the performance of each method after convergence. This shows that our method outperforms all the baselines when the labelers do not have equal knowledge about all classes. Since we only have YN responses, an ABCD model would not be appropriately trained.

\begin{itemize}
\item \textit{YN query}: We predicted $\mathbf{Z}$ via posterior inference.
\item \textit{Each labeler's ABCD simulated votes}: We asked one $k_i^j$ per object to each labeler, where $k=z_i$. This means we asked if $\mathcal{X}_i$ belongs, ``yes'' or ``no'', to what we know is the true label $z_i$. We considered these answers as ABCD votes. We obtained the classification accuracy score as the proportion of right answers.
\item \textit{Majority vote}: As a prediction, we took the majority of the labelers' ABCD simulated votes.
\item \textit{Average vote}: Represents the average of the accuracy scores of each labeler's ABCD simulated votes.
\end{itemize}

\subsection{Performance Depending on the Training Set Size - MACHO Data.}
\label{result:3}
\begin{figure}[!h]
    \centering
    \includegraphics[width=.6\linewidth]{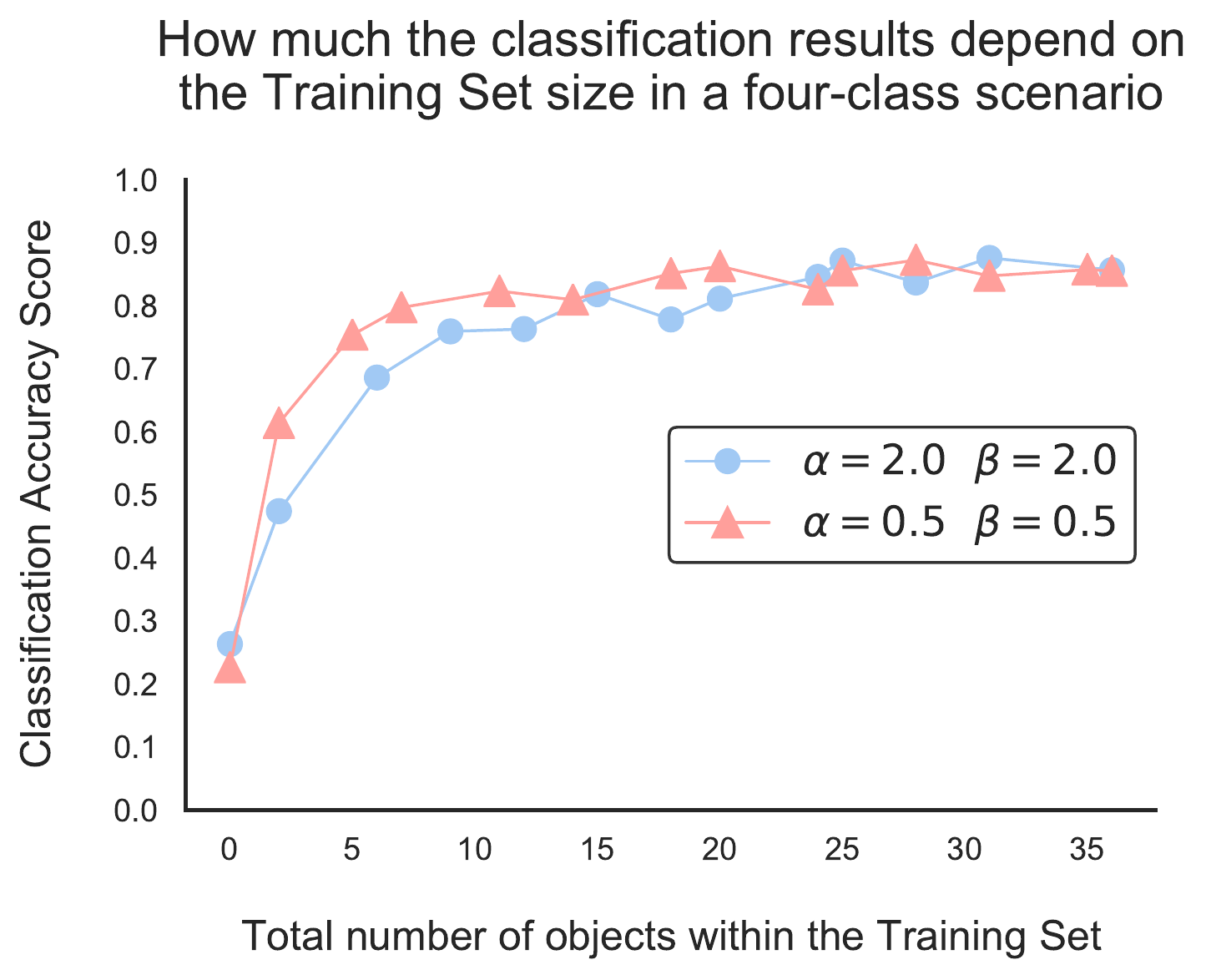}
    \caption{Classifiers voting for MACHO data. Note that increasing the training set size to about 10 instances produces an increase of $50\%$ (from $20\%$ to $70\%$) in the classification accuracy score. This shows that, after a small number of instances, the accuracy remains stable.}
    \label{fig:alpha-beta}
\end{figure}

First, we evaluated how many objects we would need to converge the $\hat{\Theta}$ estimation quickly. Second, we checked the model's sensitivity to the hyperparameters $\alpha$ and $\beta$. Figure \ref{fig:alpha-beta} shows that the learning rate grows logarithmically with the training set size. This means that by only asking about a few known objects $\hat{\mathcal{X}}_i$, the model can quickly converge to a good estimation of $\hat{\Theta}$ and $\mathrm{P(\mathbf{Z}|\mathbf{R})}$, almost independently of $\mathcal{N}$. It also shows that this model can achieve equal results with different initial hyperparameter values.

\subsection{Recovery of Credibility Matrices $\hat{\Theta}$ - MACHO Data.}
\label{result:4}
\begin{figure}[!h]
    \centering
    \includegraphics[width=.7\linewidth]{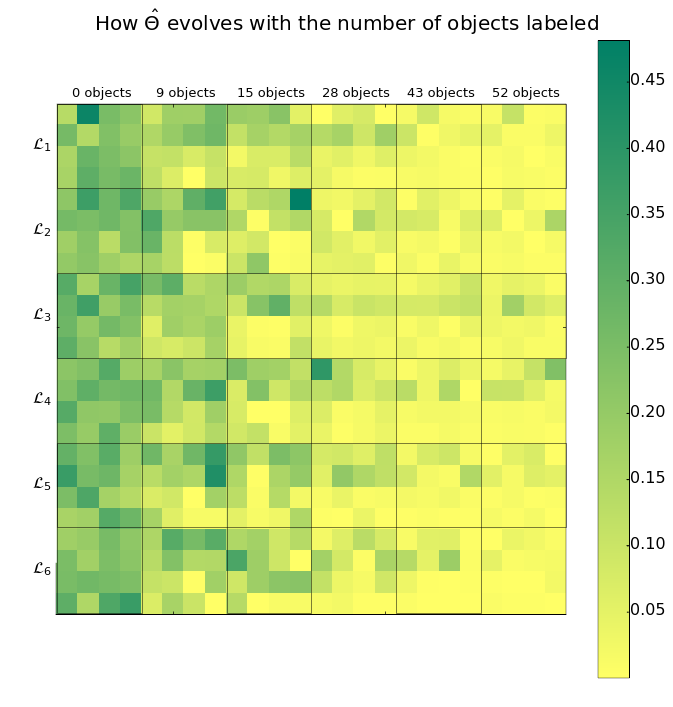}
    \caption{Classifiers voting for MACHO data MSE. MSE between each original Credibility Matrix row and its recovered $\hat{\theta}_{kk'}^{j}$ estimation with our method. 
    Figures \ref{fig:alpha-beta} and  \ref{fig:theta-hat-over-time} both show convergence by the object number 35. If we have 36 objects and 4 classes, each labeler votes for about 9 objects per class. $\mathrm{E}\left\{\rm{Random}(1, 4)\right\} = 2.5$ questions per object implies $22.5$ questions per class, which means about $5.6$ votes to estimate each $\hat{\theta}_{kk'}^j$. We can see that the convergence of $\hat{\Theta}$ depends directly on that set size.}
    \label{fig:theta-hat-over-time}
\end{figure}

The accuracy classification score and the training set size are closely related, as shown in figure \ref{fig:alpha-beta}. Figure \ref{fig:theta-hat-over-time} shows that the convergence of $\hat{\Theta}$ also depends on the training set size. Hence, if we estimate a good $\hat{\Theta}$, we can reach a higher accuracy score. Finally, the accuracy score depends on the convergence of $\hat{\Theta}$.

\subsection{Performance Simulations Depending on $\Theta$ Convergence - MACHO Data.}
\label{result:5}
\begin{figure}[!h]
    \centering
    \includegraphics[width=.8\linewidth]{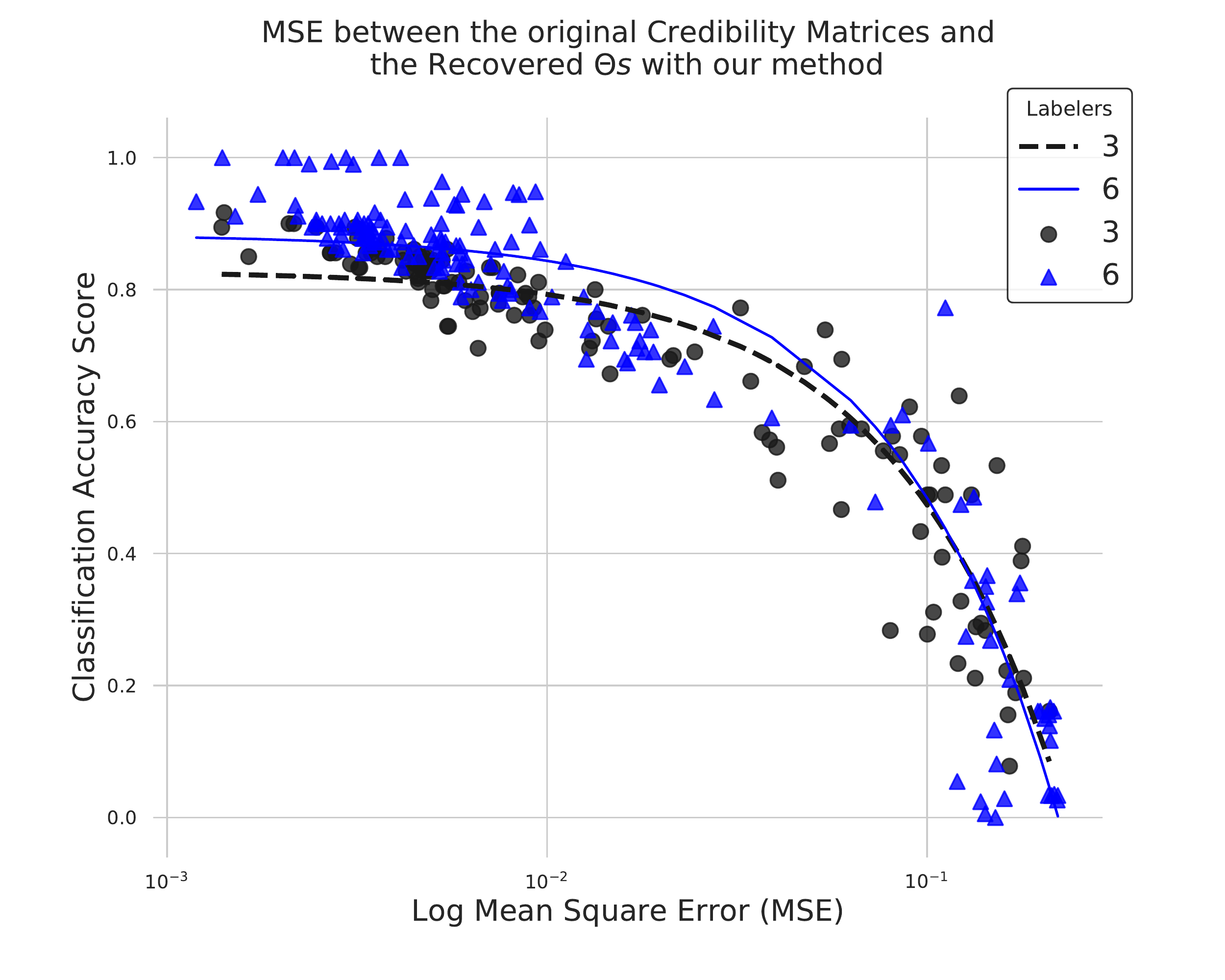}
    \caption{MSE between original Credibility Matrices and the Recovered ones. Classifiers voting for MACHO data. The error has two possible sources: i) an insufficient size of the training set, and ii) a lack of convergence in the model. In conclusion, how accurate is the estimation of $\mathrm{P(\mathbf{Z}|\mathbf{R})}$ depends on the quality of the estimation of $\Theta$.}
    \label{fig:sqerror}
\end{figure}

Figure \ref{fig:sqerror} shows that the better the model estimates the labelers' credibilities $\Theta$, the better the classification accuracy score.

\subsection{Performance Simulations - MACHO Data.}
\label{result:6}
In a four-class scenario, our method reaches the performance of the ABCD method (see figure \ref{fig:all_violins}) when we asked $\mathrm{Random}(1,4)$ YN queries per object per labeler. The implemented baseline is a Bayesian ABCD model, a Hybrid Confusion Matrix \cite{liu2012truelabel+} based on DawidSkene \cite{dawid1979maximum} plus the prior estimation stage of confusion matrices.

\begin{figure}[!h]
\centering   
	\includegraphics[width=1.1\linewidth]{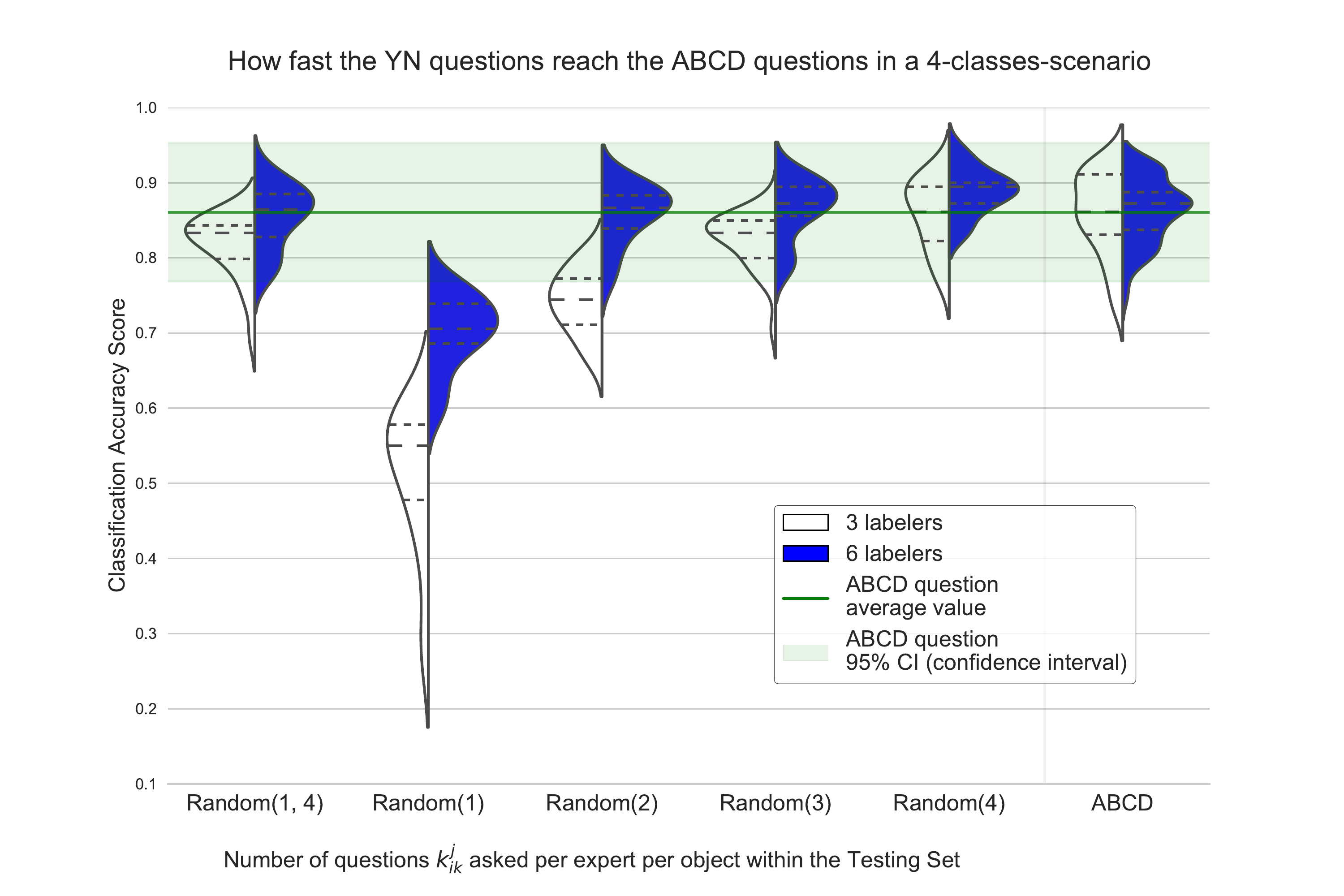} 
    \caption{Classifiers voting for MACHO data. Random($w$) means we asked each labeler for $w$ different classes $M_k$ a question $k_{ik}^j$, $w \leq \mathcal{K}$. The violin shape represents the cross validation results distribution.}
   \label{fig:all_violins}
\end{figure}

In a five-class scenario, six labelers outperformed the ABCDE model when giving responses for only four classes. This means that the labelers were not required to discern among the five classes to reach high accuracy scores. However, we found that three labelers are not enough for this scenario, since they need to respond for all five classes to reach the full question model. 

Scenarios with four and five classes showed that the YN model outperforms the ABCD method when we ask a YN question for every possible class $\mathcal{M}_k$ for every object $\mathcal{X}_i$. This indicates that each YN response is more precise or confident than each ABCD response. The difference relies on the fact that in the YN model we can ask for enough explicit information to estimate each row of the credibility matrices, while in the ABDC scenario, we cannot ask queries to evaluate specific errors between pairs of classes.

\subsection{Performance Real-World Votes MCMC vs. BBVI - Websites.}
\label{result:7}
We ran all previous simulations using the PyMC3 implementation mainly for two reasons. First, even though we used the AdaGrad \cite{ranganath2014black} algorithm to set the learning rate, this setting presents more parameter tunning than does MCMC parametrization in BBVI. Second, the PyMC3 implementation usually slightly outperformed the BBVI results. Even though we also evaluated time and memory complexity, here we present only time until complete convergence.

\subsubsection*{Time Until Complete Convergence} The experiments were performed for times of 10 minutes (PyMC3) versus 5 minutes (BBVI) for The Catalina Surveys full model running 1 chain; the times for the Animals Dataset were 14 minutes (PyMC3) versus 7 minutes (BBVI). Since both datasets were equal in siz, those times depend only on the number of labelers, 8 and 11 respectively for each dataset. The time spent is linear on the number of chains for both models. 

Given that the experiments took minutes to converge, these implementations cannot support active learning, as each step would require converging a model to estimate the next question and labeler.

\begin{figure}[!h]
    \centering
    \includegraphics[width=.87\linewidth]{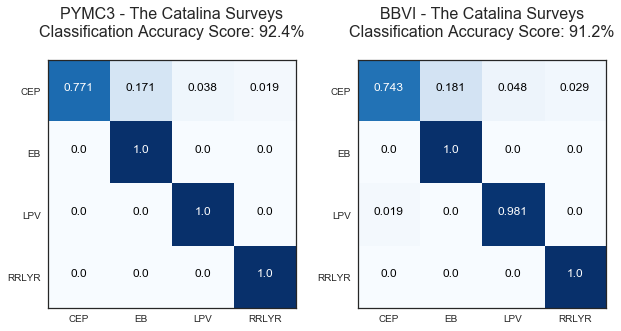}
    \caption{PyMC3 vs. BBVI. Confusion matrices for the learned models from  Real-World Data.}
    \label{fig:bbvi}
\end{figure}

The results for The Catalina Surveys are shown in figure \ref{fig:bbvi}. The figure shows that for this data, the MCMC model outperforms the BBVI implementation. For the Animals Data, both implementations have a 99.7\% accuracy score. The BBVI implementations are both parametrized equally. We found that the BBVI approach can get higher accuracy if we fine-tune each learning rate of the latent variables.

\subsection{Performance Crowd Versus Each Labeler - Websites.}
\label{result:8}
\begin{figure}[!h]
    \centering
    \includegraphics[width=.95\linewidth]{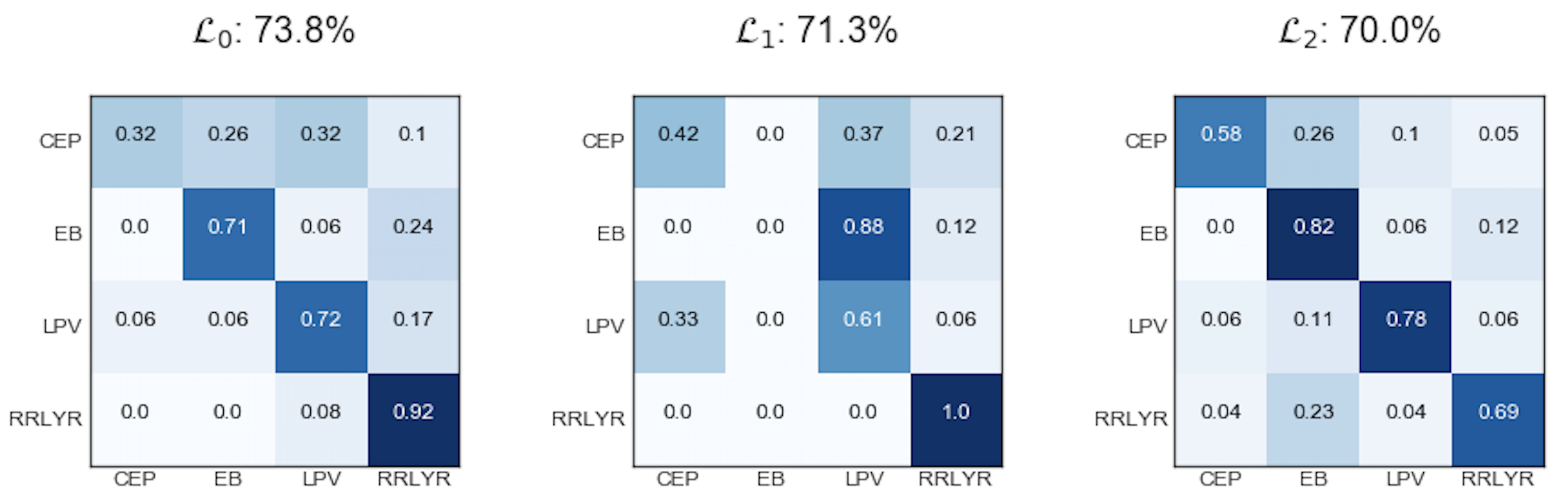}
    \caption{The Catalina Surveys contest's participants. Even though the labelers are confused, the YN model can learn how they fail. Figure \ref{fig:bbvi} shows that our model outperforms each labeler.}
    \label{fig:each-cata}
\end{figure}

To evaluate the individual performance of each labeler versus the mixture of them, we trained one YN model per labeler. Figure \ref{fig:each-cata} shows the three best individual performances in the The Catalina Surveys contest. The figure shows that our strategy effectively modeld and integrated the unreliable crowd knowledge. 

The YN strategy can control unreliable labelers mainly for two reasons. First, the Credibility stage allows the model to discover how each labeler makes mistakes and interprets the labelers' responses. Second, the mixture of labelers helps the model to converge to a correct posterior distribution of the classes by weighting them according to their credibility matrices.

The labelers' behavior for the Animals datasets is quite similar; many of them are unreliable, but the full model is more accurate than all the labelers.

\subsection{Performance Real-World Votes YN vs. ABCD - Websites.}
\label{result:9}
As we explained in section \ref{sec:data}, each labeler was presented with a series of full ABCD questions for 80 objects, for which the labelers were asked for $\rm{Random}(1, 4)$ YN queries as well. For these objects, the animals contest achieved 100\% accuracy with both strategies. For The Catalina Surveys, the YN query reached 91.2\% and the ABCD 90.0\%.

\subsection{Performance Analysis YN Question vs. ABC Question - Websites.}
\label{result:10}
\begin{figure}[h]
    \centering
    \includegraphics[width=.65\linewidth]{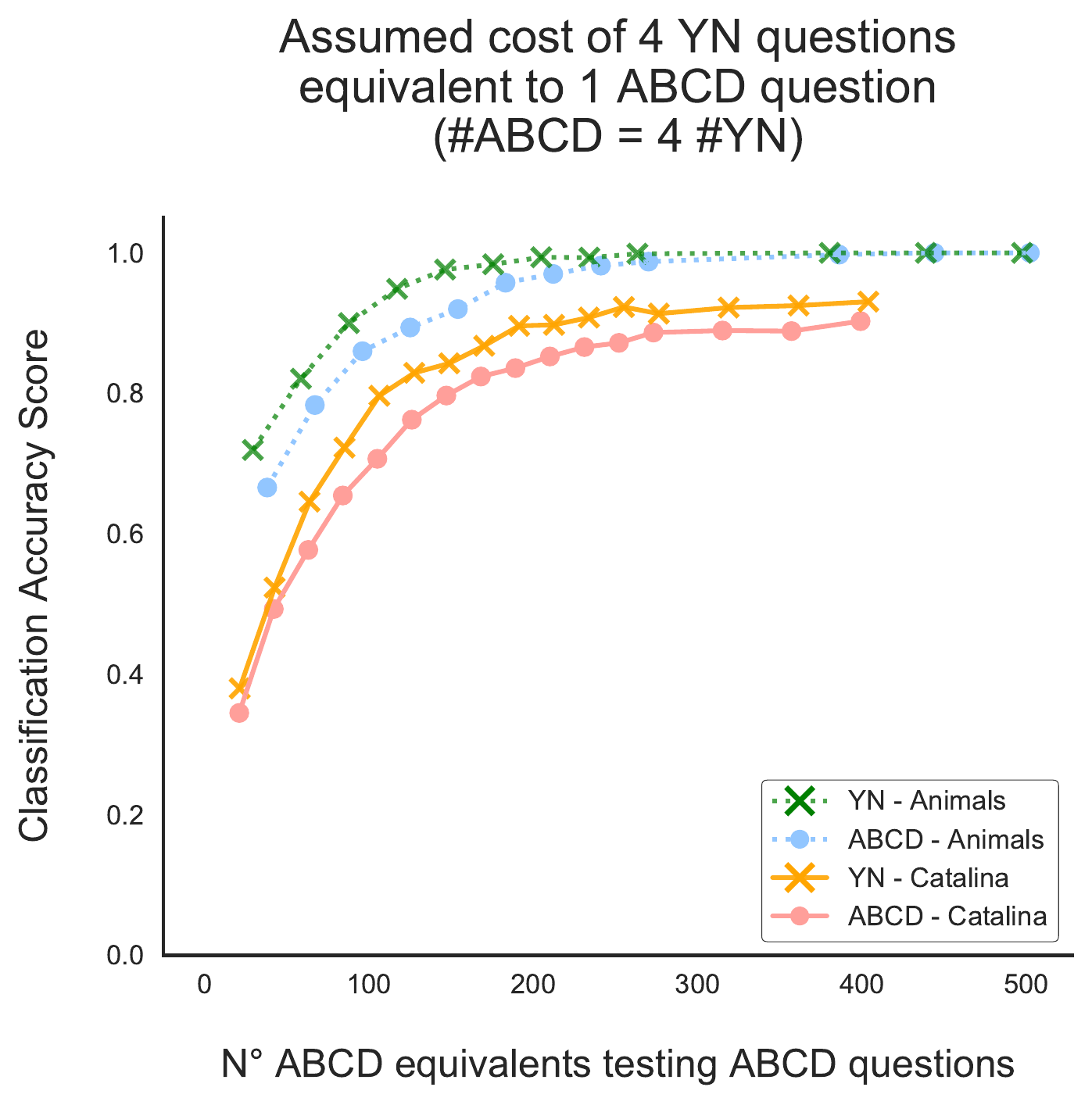}
    \caption{ABCD equivalent questions. Results from two web contests: Real-World votes on two different scenarios. The ABCD predictions were obtained from the Bayesian model described in section \ref{result:6}.}
    \label{fig:abcd-4yn}
\end{figure}

Finally, we analyzed the cost and performance of the number of YN queries versus the number of ABCD queries needed for convergence of the classification accuracy score. Although the YN query requires less expertise than the full ABCD question, the time spent on selecting an ABCD response is not proportional to the number of possible classes $\mathcal{K}$. This is shown in the websites' time records, where answering an ABCD question required less than twice the time of answering a YN question. To measure the cost, we compared how many YN queries versus how many ABCD queries are needed for the model to converge. We could assume that each ABCD query is equivalent to give $\mathcal{K}$ YN votes \cite{welinder2010online}, because each ABCD response requires the labeler to recognize the YN response for all $\mathcal{K}$ possible classes. Figure \ref{fig:abcd-4yn} shows that if 4 YN queries require as much effort as 1 ABCD question, the YN model converges faster and to a higher classification accuracy score. This occurs because the YN model can better differentiate among the possible errors, since the YN query gives specific information to estimate all the rows within the credibility matrices. As figure \ref{fig:sqerror} shows, the better the model estimates the credibility matrices, the better the classification accuracy score.

\begin{figure}[h]
    \centering
    \includegraphics[width=\linewidth]{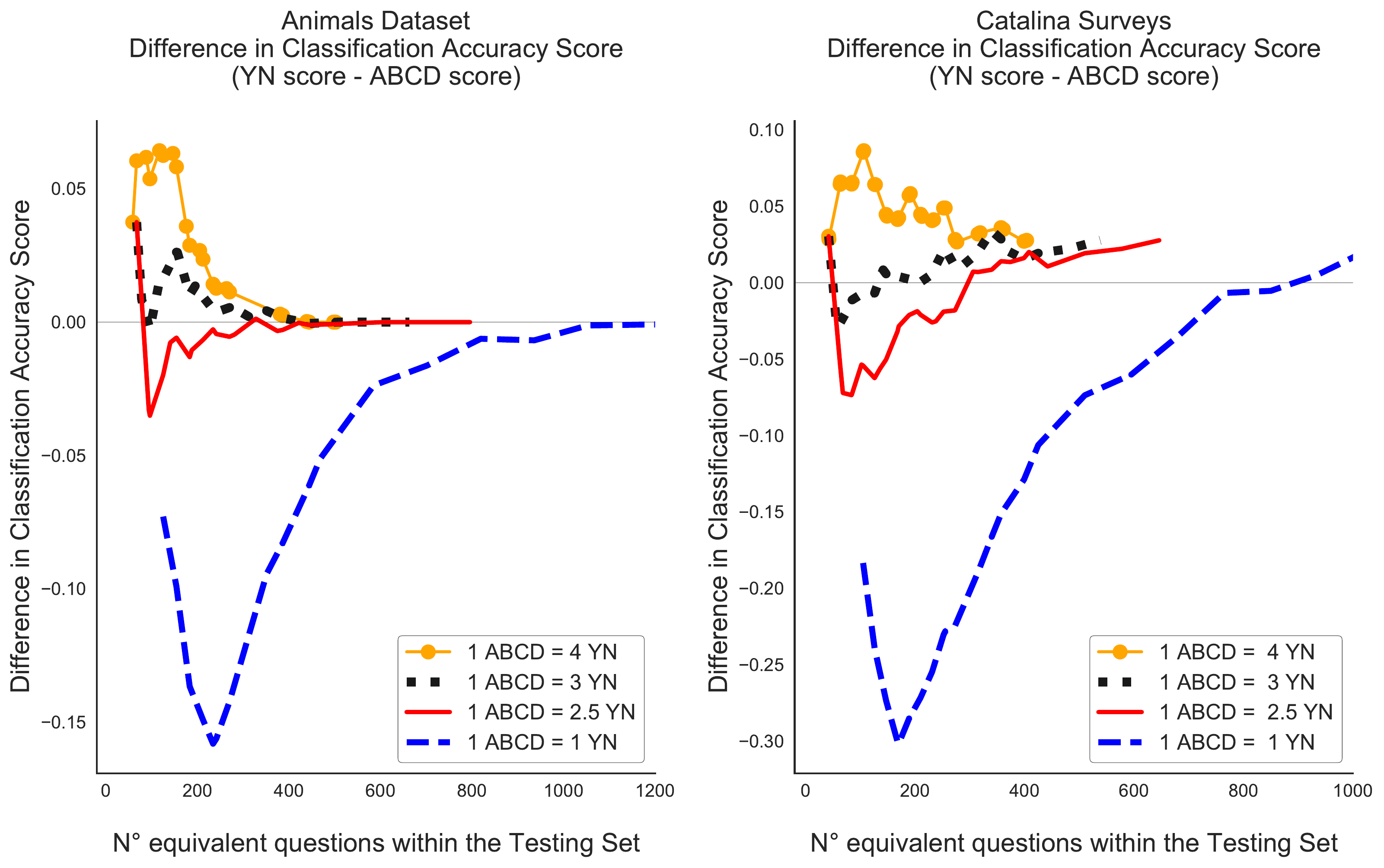}
    \caption{Difference in Classification Accuracy Scores. Results from two web contests: Real-World votes on two different scenarios. The ABCD predictions were obtained from the Bayesian model described in section \ref{result:6}.}
    \label{fig:error-difference}
\end{figure}

Despite assuming that 4 YN queries are equivalent to 1 ABCD query, figure \ref{fig:error-difference} presents an analysis of different ABCD equivalences. All ABCD predictions were obtained from the Bayesian model described in section \ref{result:6}, which was also used in figure \ref{fig:all_violins}.

The analysis in figure \ref{fig:error-difference} corresponds to how much difference exists between the classification accuracy score of the YN scenario and that of the ABCD scenario. The ``1 ABCD = 4 YN'' lines represent the differences in figure \ref{fig:abcd-4yn}, where the YN surpasses the ABCD strategy. We compared this error (axis-Y) to the number of equivalent ABCD questions asked during the labeling stage (axis-X). Figure \ref{fig:error-difference} illustrates that the YN strategy outperforms the ABCD strategy when we assumed that each ABCD query is equivalent to at least 3 YN queries. In addition, we can see that when asking an average of 2.5 questions per object and labeler, the YN model reached the ABCD's performance quickly. Furthermore, when we assume that each YN question is equivalent in cost to one ABCD question, at some point the YN reaches or outperforms the ABCD's performance.

\subsection{Cognitive Cost Analysis YN Question vs. ABC Question - Websites.}
\label{result:11}

\begin{figure}[!h]
    \centering
    \includegraphics[width=.65\linewidth]{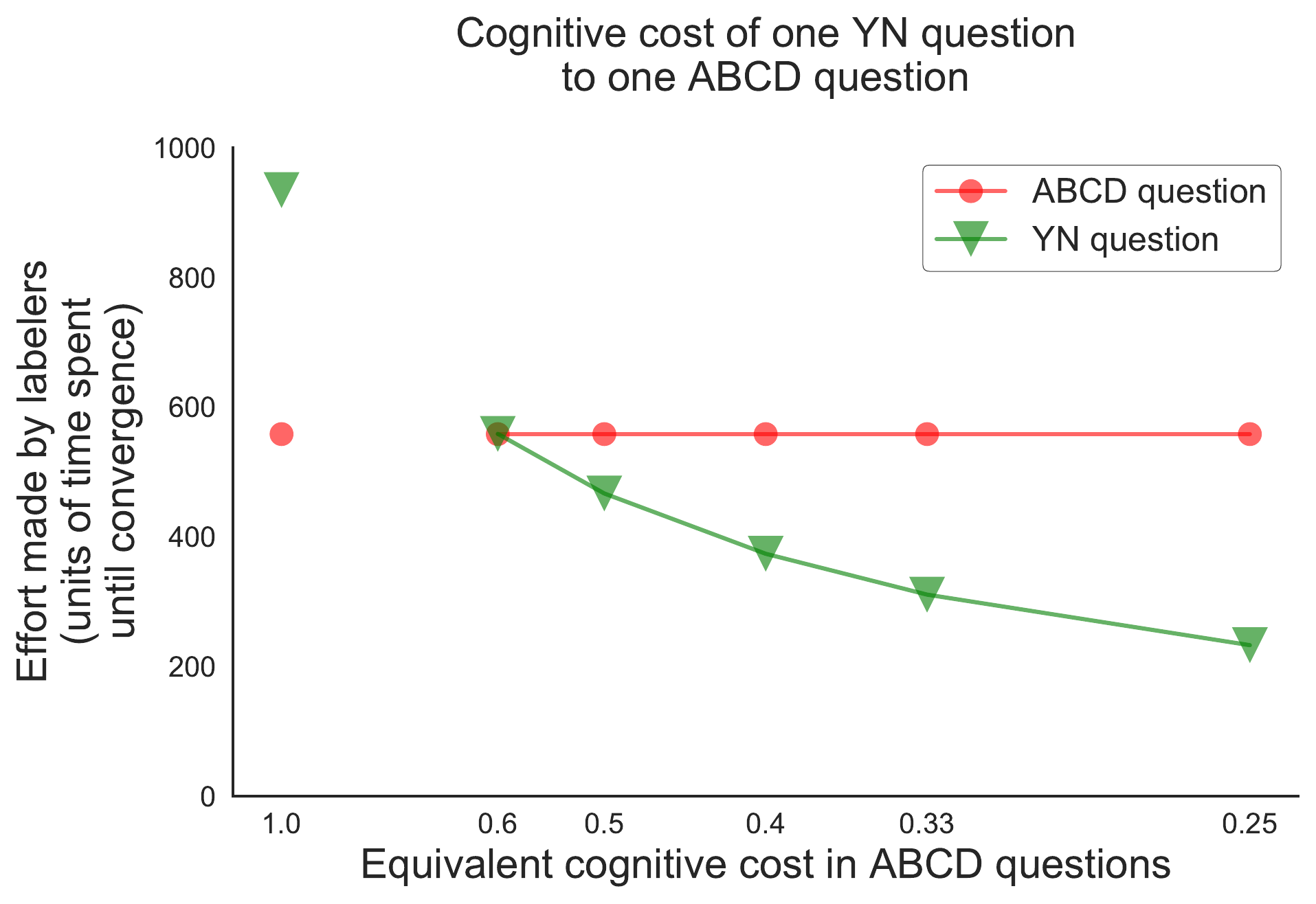}
        \caption{Results from web contests: Real-World votes on two different scenarios. ABCD predictions are from the Bayesian model described in section \ref{result:6}. The y-axis values were taken from the website scenarios. The times marked at cost 1.0 are empirical data, and any other point is proportional to the assumed cognitive effort.}
    \label{fig:cognitive-costs}
\end{figure}

The amount of cognitive effort made by annotators depends on factors like the information available or the number of classes. Since we cannot evaluate all possible scenarios objectively, we show the assessment of different costs in a four-class scenario in figure \ref{fig:cognitive-costs}. Figure \ref{fig:cognitive-costs} illustrates that assuming that each ABCD query is equivalent to one YN query, the model is not convenient regarding time spent. However, when the cognitive cost of a YN query is less than half that of an ABCD query, the effort made by annotators to converge the model is less than the effort required when they are asked for ABCD queries. Overall, we can see that if the cognitive cost for a YN query is less than 0.6 times that for an ABCD query, the YN strategy reduces the total effort.

\section{Conclusion}
\label{sec:conclusions}
We developed a new model for crowdsourcing with ``yes'' or ``no'' type queries that can be applied to any context. The YN model obtains comparable results with models that ask full questions to labelers. The reduction of labelers' efforts depends on how much cognitively easier it is to respond to a YN versus an ABCD question. Furthermore, our model convergences more quickly without sacrificing accuracy. We could also see that in cases where most labelers are unreliable, the YN model was able to capture the right posterior of the classes by taking advantage of crowds. 

As a future work, the model could capture variations in expertise over time. Also, here we randomly selected an object along with a class; this election could be optimized using an active learning approach or by understanding the biases produced by the order in which the pairs of objects and questions are presented to the labelers.

\section*{Acknowledgements}
\scriptsize
Our work was supported in part by the CSS survey, which is funded by the National Aeronautics and Space Administration under Grant No. NNG05GF22G issued through the Science Mission Directorate Near-Earth Objects Observations Program. We would also like to thank the anonymous reviewers whose comments greatly improved this manuscript.

\newpage
\section*{\Large Supplementary Material\\}
\normalsize

\section{Background Theory}
\label{sec:background}
This section describes the main theory behind this work. We based this discussion mainly on \cite{blei2017variational} and \cite{murphy2012machine}.

\subsection{Probabilistic Graphical Models}
\begin{figure}[!h]
    \centering
    \includegraphics[width=3in]{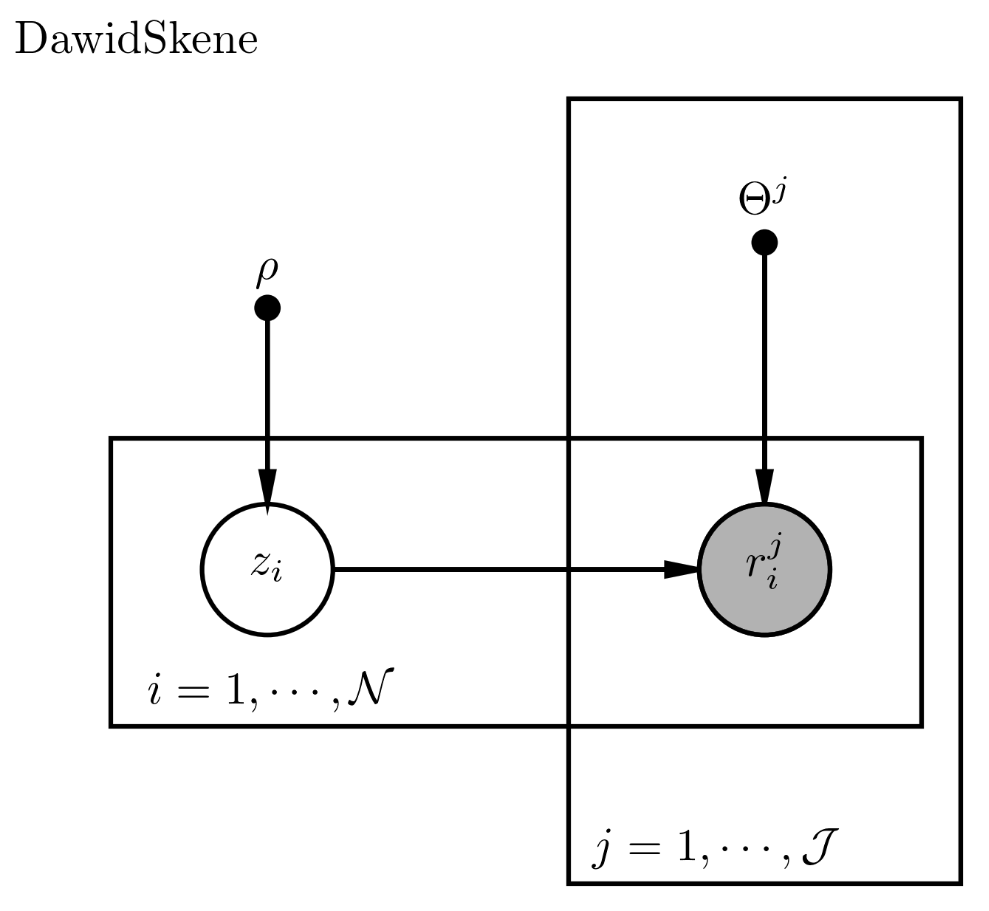}
    \caption{The DawidSkene model represented as a PGM in plate notation.}
    \label{fig:DawidSkene}
\end{figure}

We represented the joint distribution of the proposed method with a probabilistic graphical model (PGM) [14, 29]. A PGM is a graph-based representation for compactly encoding a complex distribution over a high-dimensional space. For example, figure \ref{fig:DawidSkene} illustrates the elemental DawidSkene [4] distribution for a crowdsourcing classification scenario. The circles represent random variables, observed variables are gray circles, and the points represent hyperparameters. When a set of variables shares the same probability distribution, we can use the ``plate'' notation, which stacks identical objects in a rectangle. In that case, the plates' dimensions are written in capital letters within the rectangles.

In the PGM shown in figure \ref{fig:DawidSkene}, $\mathcal{N}$ is the number of instances to be labeled and $\mathcal{J}$ is the number of labelers, where $i \in \{1,..\mathcal{N}\}$ and $j \in \{1,..\mathcal{J}\}$. In the DawidSkene model, $\rho$ is the initial parameter for the distribution over the hidden labels $\mathbf{Z}$, where $z_i$ is the predicted label for object $\mathcal{X}_i$. In that scenario, $r_{i}^{j}$ represents the class given by labeler $L_j$ to object $\mathcal{X}_i$, whose confusion matrix is $\Theta^j$. In this case, if each $\Theta^j$ is a random variable instead of a hyperparameter, $\mathbf{Z}$ and $\Theta$ will be conditionally dependent given all the labelers' votes $\mathbf{R}$ due to the graph structure. Following the notation from [17], in that model definition the variable distributions are:
\begin{equation}
z_i \sim \mathrm{Multinomial}(\rho)
\end{equation}
\begin{equation}
r_i^j \sim \mathrm{Multinomial}(\Theta^j(z_i,:))
\end{equation}

This structure allows inferring a compact representation of the explicit joint distribution. To get the posterior distribution, we can either use sampling-based methods or variational inference. In this work, we address the proposed probabilistic model solution with approximate inference. In the following subsections, we explain two approaches to infer the posterior target distribution by approximating a distribution: Markov chain Monte Carlo (MCMC) and Variational Inference (VI).

\subsection{Markov Chain Monte Carlo}
MCMC [10, 7] is the most popular method for sampling when simple Monte Carlo methods do not work well in high-dimensional spaces. The key idea is to build a Markov chain on the state space $\mathcal{Z}$ where the stationary distribution is the target, for instance, a posterior distribution $\mathrm{p}(z | x)$, where $x$ is observed data. MCMC performs a random sampling walk on the $\mathcal{Z}$ space, where the time spent in each state $z$ is proportional to the target distribution. The samples allow approximating $\mathrm{p}(z | x)$.

MCMC approaches Bayesian inference with developments as the Gibbs sampler [9]. The key idea behind Gibbs sampling is to turn the sampling among the variables. In each turn, the sampler conditions a new variable sample $s$ on the recent values of the rest of the distributions in the model. 
Suppose we want to infer $\mathrm{p}(z_1, z_2)$. In each iteration, we would turn the samples iteratively:  $z_1^{s+1} \sim \mathrm{p}(z_1 | z_2^{s})$ and  $z_2^{s+1} \sim \mathrm{p}(z_2| z_1^{s})$. 

\subsubsection*{No-U-Turn Hamiltonian Monte Carlo (NUTS)}
To avoid the random walk and converge the sampling more quickly than with simple MCMC, we used NUTS [9], an MCMC algorithm based on a Hamiltonian Monte Carlo sampler (HMC). As an advantage, NUTS uses an informed walk and avoids the random walk by using a recursive algorithm to obtain a set of candidate points widely spread over the target distribution. Furthermore, NUTS stops when the recursion starts to go back to trace the dropped steps again. Nevertheless, HMC requires computing the gradient of the log-posterior to inform the walk, which can be difficult.

Using NUTS does not require establishing the step size and the number of steps to converge, compared to what a simple MCMC or HMC sampler does. Setting those parameters would require preliminary runs and some expertise. This sampling stops when drawing more samples no longer increases the distance between the proposal $\tilde{z}$ and the initial values of $z$.

Even though MCMC algorithms can be very slow when working with large datasets or very complex models, they asymptotically draw exact samples from the target density \cite{robert2004monte}. Under these heavy computational settings, we can use variational inference (VI) as an approximation to the target distribution. VI does not guarantee finding the density distribution, it only finds a close distribution; however, it is usually faster than simple MCMC.

\subsection{Variational Inference}
Variational inference (VI) \cite{jordan1999introduction} proposes a solution to the problem of posterior inference. VI selects an approximation $\mathrm{q}(z)$ from some tractable family and then tries to make this $\mathrm{q}(z)$ as close as possible to the true posterior $\mathrm{p^{*}}(z) \overset{\Delta}{=} \mathrm{p}(z | x)$. The VI approach reduces this approximation to an optimization problem: the minimization of the KL divergence \cite{kullback1951information} from $\mathrm{q}$ to $\mathrm{p}^{*}$.

The KL divergence is a measure of the dissimilarity of two probability distributions, $\mathrm{p}^{*}$ and q. Given that the forward KL divergence $\mathbb{KL}(\text{p}^{*} || \text{q})$ includes taking expectations over the intractable $\mathrm{p^{*}}(z)$, a natural alternative is the reverse KL divergence $\mathbb{KL}(\text{q} || \text{p}^{*})$, defined in (\ref{eq:kl}).
\begin{equation}
\mathbb{KL}(\text{q} || \text{p}^{*}) = - \int  \text{q}(z)   \log  \frac{\text{q}(z)}{\text{p}^{*}(z)}   dz
\label{eq:kl}
\end{equation}
\subsection{The Evidence Lower Bound}

Variational inference minimizes the KL divergence from $\mathrm{q}$ to $\mathrm{p}^{*}$. It can be shown to be equivalent to maximize the lower bound (ELBO) on the log-evidence $\mathrm{log\ p}(x)$. The ELBO is equivalent to the negative KL divergence plus a constant, as we show in the following definitions.

Assume ${x}$ is the observations, ${z}$ the latent variables, and $\lambda$ the free parameters of $\mathrm{q}({z}|\lambda)$. We want to approximate $\mathrm{p}({z}|{x})$ by setting $\lambda$ such that the KL divergence is minimum. In this case, we can rewrite (\ref{eq:kl}) and expand the conditional in (\ref{eq:kl1}).
\begin{equation}
\mathbb{KL}(\mathrm{q} || \mathrm{p}^{*}) =  \mathbb{E}_{\mathrm{q}}[\mathrm{log\ q}(z|\lambda)] -  \mathbb{E}_{\mathrm{q}}[\mathrm{log\ p}(z,x)] - \mathrm{log\ p}(x)
\label{eq:kl1}
\end{equation}

Therefore, the minimization of the KL in (\ref{eq:elbo0}) is equivalent to maximizing the ELBO:
\begin{equation}
\mathcal{L}(\mathrm{q}) = \mathbb{E}_{\mathrm{q}}[ \mathrm{log\ p} (z,x) - \mathrm{log\ q} (z|\lambda) ]
\label{eq:elbo0}
\end{equation}

\subsection{Mean Field Inference}
Optimization over a given family of distributions is determined by the complexity of the family. This optimization can be difficult to optimize when a complex family is used. To keep  the variational inference approach simple, \cite{opper2001advanced} proposes to use the mean field approximation. This approach assumes that the posterior can be approximated by a fully factorized $\mathrm{q}$, where each factor is an independent mean field variational distribution, as is defined in (\ref{eq:mean-field-def}).
\begin{equation}
\mathrm{q}(z) = \prod_{i=1}^{m} \mathrm{q}_i(z_i)
\label{eq:mean-field-def}
\end{equation}

The goal is to solve the optimization in (\ref{eq:opti-problem-mf}) over the parameters of each marginal distribution q.
\begin{equation}
\underset{\lambda_1,...,\lambda_m}{\mathrm{min}}\ \mathbb{KL} (\mathrm{q} || \mathrm{p}^{*})
\label{eq:opti-problem-mf}
\end{equation}

\subsection{Stochastic Variational Inference}
Common posterior inference algorithms do not easily scale to work with high amounts of data. Furthermore, several algorithms are very computationally expensive because they require passing through the full dataset in each iteration. Under these settings, stochastic variational inference (SVI) \cite{hoffman2013stochastic} approximates the posterior distribution by computing and following its gradient in each iteration over subsamples of data. SVI iteratively takes samples from the full data, computes its optimal local parameters, and finally updates the global parameters.

SVI solves the ELBO optimization by using the natural gradient \cite{amari1998natural} in a stochastic optimization algorithm. This optimization consists of estimating a noisy but cheap-to-compute gradient to reach the target distribution.  

\subsection{Black Box Variational Inference}
The BBVI [20] avoids any model-specific derivations. Black Box VI proposes stochastically maximizing the ELBO using noisy estimates of its gradient. The estimator of this gradient is computed using samples from the variational posterior. Then, we need to write the gradient of the ELBO (\ref{eq:elbo0}) in (\ref{eq:gradient0}).
\begin{equation}
\nabla_{\lambda} \mathcal{L} = \mathbb{E}_\mathrm{q} [ \nabla_{\lambda} \mathrm{log\ q}(z|\lambda) ( \mathrm{log\ p} (z, x) - \mathrm{log\ q} (z|\lambda) )  ]
\label{eq:gradient0}
\end{equation}

Using this equation, we can compute the noisy unbiased gradient of the ELBO sampling the variational distribution with Monte Carlo, as shown in equation (\ref{eq:app-elbo0}), where $S$ is the number of samples we take from each distribution to be estimated.
\begin{equation}
\nabla_{\lambda} \mathcal{L} \approx \frac{1}{S}
\sum_{s=1}^{S}{
\nabla_{\lambda} \mathrm{log\ q}(z_s|\lambda) ( \mathrm{log\ p} (z_s,x) - \mathrm{log\ q} (z_s|\lambda) ) }
\label{eq:app-elbo0}
\end{equation}
where, 
\begin{equation}
 z_s \sim \mathrm{q}(z|\lambda)
\end{equation}

For estimating the approximating q distribution, in BBVI the variational distributions $\mathrm{q}(z_i)$ are mean field factors with free variational parameters $\lambda_i$, for each index $i$ (see (\ref{eq:mean-field-def})). In appendix \ref{ap:bbvi}, we show how to apply this method to the proposed model.

\section{Inference Schema}

\begin{figure}[!t]
    \centering
    \includegraphics[width=\linewidth]{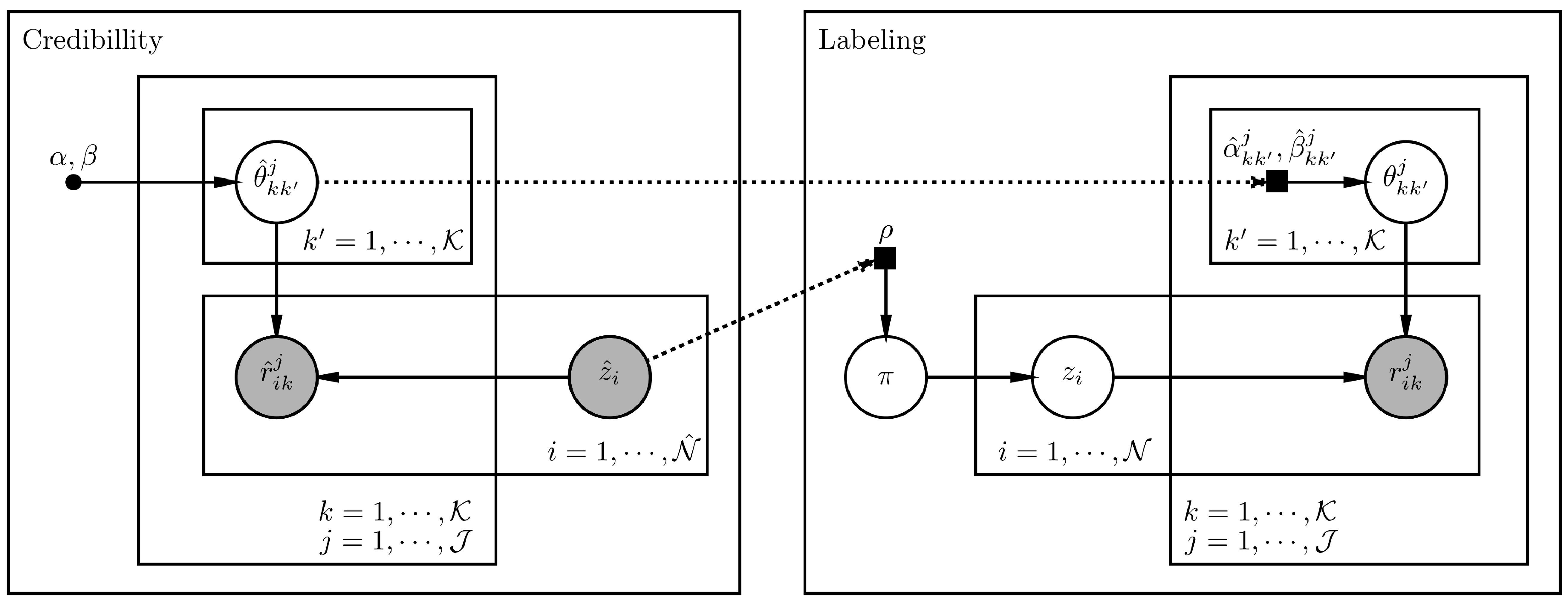}
    \caption{Proposed inference scheme for the YN model. Dashed lines represent prior parameters estimated in the \texttt{Credibility} stage. These parameters were then used as input to the \texttt{Labeling} stage.}
    \label{fig:inference-scheme}
\end{figure}

As stated in the paper, the proposed inference scheme works in two stages. The PGM in \ref{fig:inference-scheme} shows both stages.

\section{Complementary Results}

\subsection{Convergence Simulations - Synthetic Data.}

\begin{figure}[!h]
    \centering
    \includegraphics[width=\linewidth]{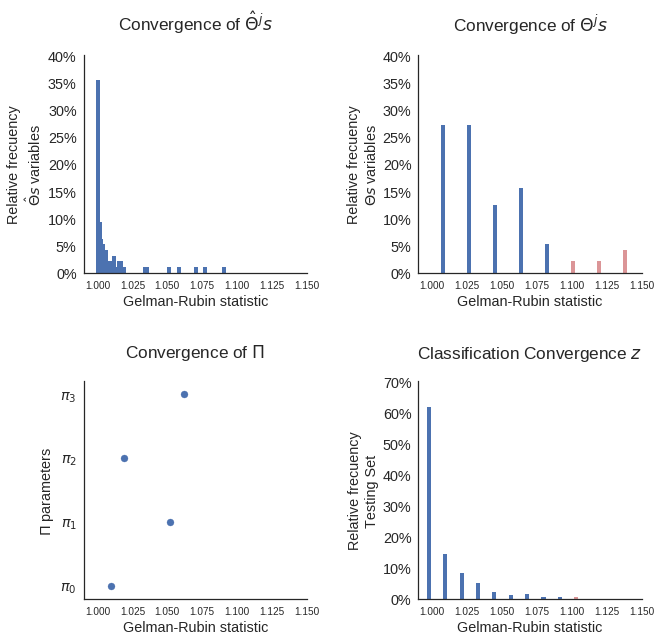}
    \caption[Accuracy score convergence]{Accuracy score convergence. It is possible to see that $\Theta$ has more variance than any other variable, and some of the $\theta$s have not completely converged. We cannot conclude that the model has not converged, we can only say that one of the chains has not converged. In practice, that minimum percentage does not condition the full model.}
    \label{fig:convergence}
\end{figure}

To check for convergence of the full model, we analyzed each variable convergence. The convergence diagnostics for our random variables was based on the Gelman-Rubin statistic [7]. To try this diagnostic, we needed multiple chains to compare the similarity between them. Our experiments were based on 10 chains each. When the Gelman-Rubin ratio (potential scale reduction factor) is less than 1.1, it is possible to conclude that the estimation has converged. Figure \ref{fig:convergence} presents the potential scale reduction factors for all the estimated variables. According to this figure, there is no disagreement on whether each $z_i$ converges.

\subsection{Performance Simulations - MACHO Data.} Figure \ref{fig:all_violins} shows the results for the experiment in subsection \textbf{7.6} in a five-class scenario. We can see that when all classes were asked per object per labeler, the YN model outperformed the ABCD strategy. However, three labelers are not enough for this scenario because the only way they reached the ABCDE performance (five classes implies ABCDE) was when we asked them about all five classes. In this five-class scenario, six labelers outperformed the ABCDE model when giving responses for only four classes. This means that the labelers were not required to discern among the five classes to reach a high accuracy score.

\begin{figure}[!h]
	\includegraphics[width=\linewidth]{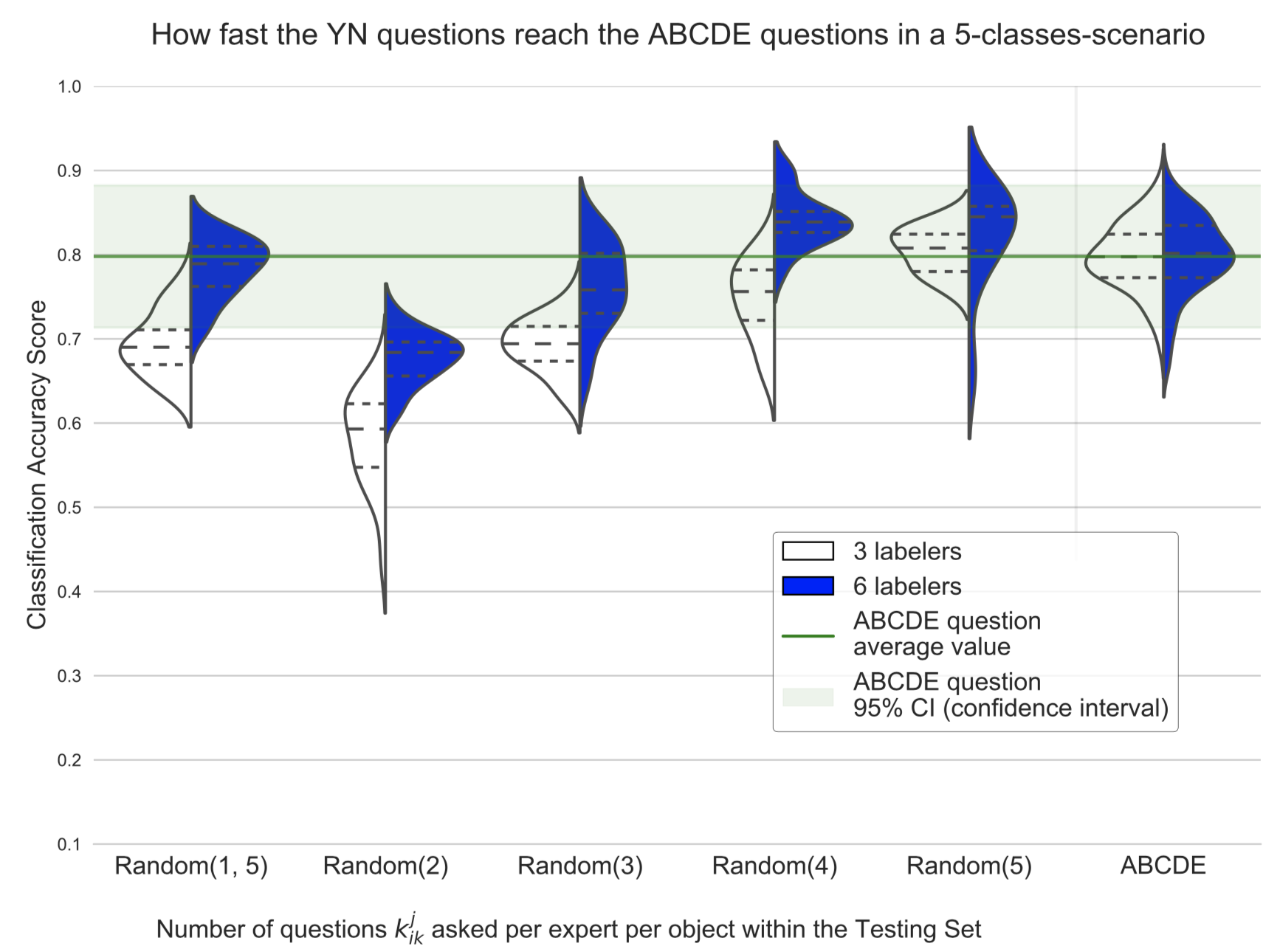} 
    \caption{Classifiers voting for MACHO data. Random($w$) means we asked each labeler for $w$ different classes $M_k$ a question $k_{ik}^j$, $w \leq \mathcal{K}$. The violin shape represents the cross validation results distribution.}
   \label{fig:all_violins}
\end{figure}

\subsection{Performance Real-World Votes MCMC vs. BBVI on Websites Results}
\label{result:7}
We developed all the previous simulations using the PyMC3 implementation mainly for two reasons. First, even though we used the AdaGrad [20] algorithm to set the learning rate, this setting presents more parameters tuning than the MCMC parametrization. Second, the results were usually slightly outperformed by NUTS.

\subsubsection*{Iterations Until Convergence} As we said before, PyMC3 needs about 3000 iterations until convergence when running one chain. BBVI needs only 4 iterations, but each iteration implies estimating the gradient of each latent variable, which means taking samples from the variational approximation distribution of every variable. This estimation converges at 3072 total samples.

\subsubsection*{Time and Memory Complexity} The model has $\mathcal{J}\times \mathcal{K}\times \mathcal{K}\times 2$ and $\mathcal{J}\times \mathcal{K}\times \mathcal{K}\times 2 + \mathcal{N}\times \mathcal{K} + \mathcal{K}$ parameters to estimate, respectively, in each stage. If we assume that always $\mathcal{J}\times \mathcal{K} < \mathcal{N}$, this model is $\Omega(\mathcal{N}\times \mathcal{K})$. Both implementation require samples. The memory complexity for the PyMC3 model is $\mathcal{O}(\mathcal{N}\times \mathcal{K}\times NumOfSamplesMCMC)$ and for the BBVI is $\mathcal{O}(\mathcal{N}\times \mathcal{K}\times NumOfSamplesBBVI)$. When the number of samples remains constant, as in this work, the complexity is $\mathcal{O}(\mathcal{N} \times \mathcal{K})$. Both time complexities are equivalent.

\section{Derivation Black Box Inference Equations.}
\label{ap:bbvi}
The BBVI minimizes the KL divergence from an approximating distribution q to the true p posterior. Lets say $x$ is the observations, $z$ the latent variables, and $\lambda$ the free parameters of $\mathrm{q}(z|\lambda)$. We want to approximate $\mathrm{p}(z|x)$ by setting $\lambda$. This optimization is equivalent to maximizing the ELBO in (\ref{eq:elbo0}):

$$\mathcal{L}(\lambda) = \mathbb{E}_{\mathrm{q}}[ \mathrm{log\ p} (z,x) - \mathrm{log\ q} (z|\lambda) ] $$

BBVI proposes stochastically maximizing the ELBO using noisy estimates of its gradient. The estimator of this gradient is computed using samples from the variational posterior. This require writing the gradient of the ELBO as in (\ref{eq:gradient0}):
$$\nabla_{\lambda} \mathcal{L} = \mathbb{E}_\mathrm{q} [ \nabla_{\lambda} \mathrm{log\ q}(z|\lambda) ( \mathrm{log\ p} (z,x) - \mathrm{log\ q} (z|\lambda) )  ]$$

Using (\ref{eq:gradient0}), we can compute the noisy unbiased gradient of the ELBO sampling the variational distribution with Monte Carlo, as shown in (\ref{eq:app-elbo0}), where $S$ is the number of samples taken from each distribution to be estimated:
$$\nabla_{\lambda} \mathcal{L} \approx \frac{1}{S}
\sum_{s=1}^{S}{
\nabla_{\lambda} \mathrm{log\ q}(z_s|\lambda) ( \mathrm{log\ p} (z_s,x) - \mathrm{log\ q} (z_s|\lambda) ) }$$
$$\mathrm{Where,}\ z_s \sim \mathrm{q}(z|\lambda)$$

Then $\lambda$ is set at each iteration $t$ as:
$$ \lambda_t = \lambda_{t-1} + \rho \nabla_{\lambda} \mathcal{L} $$

Where the learning rate $\rho$ can be fine-tuned as a global rate for all $\lambda$s or as a unique rate per $\lambda_s$.

To estimate the approximating q distribution, BBVI uses the mean field theory. Then we define the approximating distribution q as in (\ref{eq:mean-field-def}):
$$\mathrm{q}(z) = \prod_{i}^{m} \mathrm{q}_i(z_i)$$

The variational mean field distributions q from (\ref{eq:mean-field-def}) in the Credibility Estimation (first stage) of the YN model are found in (\ref{eq:t-hat}). Their free variational parameters to estimate are in (\ref{eq:t-hat-parameters}).
\begin{equation}
\mathrm{q}(\hat{\theta}_{kk'}^{j} ) \sim \mathrm{Beta}(\hat{\alpha}_{kk'}^{j},\hat{\beta}_{kk'}^{j})\ \forall\ jkk'
\label{eq:t-hat}
\end{equation}
\begin{equation}
\hat{\Theta}:\ (\hat{\alpha}_{kk'}^j,\hat{\beta}_{kk'}^j)\ \forall\ \hat{\theta}_{kk'}^j
\label{eq:t-hat-parameters}
\end{equation}

For the Labeling part (second stage) of the proposed model, the mean field distributions q from (\ref{eq:mean-field-def}) are defined in (\ref{eq:q-theta}), (\ref{eq:q-z}), and (\ref{eq:q-pi}).
\begin{equation}
\mathrm{q}(\theta_{kk'}^{j} ) \sim \mathrm{Beta}(\alpha_{kk'}^{j},\beta_{kk'}^{j})\ \forall\ jkk'
\label{eq:q-theta}
\end{equation}
\begin{equation}
\mathrm{q}(z_i) \sim \mathrm{Categorical} (\mathbf{p}_i)\ \forall\ i
\label{eq:q-z}
\end{equation}
\begin{equation}
\mathrm{q}(\pi) \sim \mathrm{Dirichlet}(\mathbf{d})
\label{eq:q-pi}
\end{equation}

Their free variational parameters to estimate are defined in (\ref{eq:params-q-theta}), (\ref{eq:params-q-z}), and (\ref{eq:params-q-pi}) respectively.
\begin{equation}
\Theta:\ (\alpha_{kk'}^j,\beta_{kk'}^j)\ \forall\ \theta_{kk'}^j
\label{eq:params-q-theta}
\end{equation}
\begin{equation}
\mathbf{z}:\ (p_{ik}\ \forall k\in K)\ \forall\ z_i
\label{eq:params-q-z}
\end{equation}
\begin{equation}
\pi:\ (d_{k}\ \forall k\in K)
\label{eq:params-q-pi}
\end{equation}

As shown in (\ref{eq:gradient0}), to maximize the ELBO, we need the expectations under q. Given that we prefer to avoid the derivation for the YN model joint distribution, we used the black box method by approximating the gradient of the ELBO as defined in (\ref{eq:app-elbo0}).

To apply this method to our model, we needed to write the needed functions for both the Credibility stage and the Labeling stage. In this appendix, we show only the derivation for that second stage (the gradients for the training part are a simplification of the presented derivations). 

\subsection{Labeling Parameters Estimation} The joint distribution to be inferred is:
\begin{multline}\mathrm{log\ p}(r_{ik}^j, z_i, \theta_{z_{i}k}^j, \pi) = \\
\mathrm{log\ p}(\theta_{z_{i}k}^j | \alpha_0,  \beta_0) + \\
\sum_{i=1}^{\mathcal{N}} \left\{ \mathrm{log\ p}(z_i | \theta_{z_{i}k}^j , \pi) + \mathrm{log\ p}(r_{ik}^j | z_i,  \theta_{z_{i}k}^j) \right\} 
\end{multline}

First, for each variable, we defined the log probability of all distributions containing the free parameters in order to obtain the mean field $q$. The priors are:
\begin{equation}\mathrm{log\ p}(\theta_{kk'}^j | \alpha_0,  \beta_0) = \mathrm{log\ Beta}(\theta_{kk'}^j | \alpha_0,  \beta_0)\end{equation}
\begin{equation}
\mathrm{log\ p}(z_i | \theta_{kk'}^j, \pi ) = \mathrm{log\ Categorical}(z_i | \pi)\end{equation}
\begin{equation}
\mathrm{log\ p}(\pi | \rho ) = \mathrm{log\ Dirichlet}( \pi | \rho)\end{equation}
\begin{multline}
\mathrm{log\ p}(r_{ik}^j | z_i, \theta_{z_{i}k}^j) = \\ \mathrm{log} \  \left\{(\theta_{z_{i}k}^j)^{r_{ik}^j[0]} \times   (1-\theta_{z_{i}k}^j)^{r_{ik}^j[1]} \right\} 
\end{multline}

Then, we wrote those log probabilities to estimate the gradient with respect to the variational parameters:

{
\begin{multline}
\mathrm{log\ q}(\theta_{kk'}^j | \alpha_{kk'}^j, \beta_{kk'}^j) = \mathrm{log\ Beta}(\theta_{kk'}^j | \alpha_{kk'}^j, \beta_{kk'}^j) =\\
\frac{\Gamma(\alpha_{kk'}^j + \beta_{kk'}^j)}{\Gamma(\alpha_{kk'}^j)\Gamma(\beta_{kk'}^j)} \times (\theta_{kk'}^j)^{\alpha_{kk'}^j-1} \times (1 - \theta_{kk'}^j)^{\beta_{kk'}^j-1}
\end{multline}}

\begin{equation}
\mathrm{log\ q}(z_i | \mathbf{p}_{i}) = \mathrm{log\ Categorical}(z_i | \mathbf{p}_i) = \sum_{k=1}^{\mathcal{K}} \left\{ z_ik \times \mathrm{log}\ p_{ik} \right\}
\end{equation}
\begin{equation}
\mathrm{log\ q}(\pi | \mathbf{d} ) = \mathrm{log\ Dirichlet}( \pi | \mathbf{d}) = \sum_{k=1}^{\mathcal{K}} \left\{ (d_k-1) \times \mathrm{log}\ \pi_k \right\}
\end{equation}

Finally, we wrote the gradients for each parameter to be estimated, where $\Psi(x) = \frac{d \Gamma(x)}{dx}$:
\begin{multline}
\nabla_{\alpha_{kk'}^j} \mathrm{log\ q}(\theta_{kk'}^j | \alpha_{kk'}^j, \beta_{kk'}^j) = \\ \mathrm{log}\ \theta_{kk'}^j  + \Psi( \alpha_{kk'}^j + \beta_{kk'}^j) - \Psi( \alpha_{kk'}^j )
\end{multline}
\begin{multline}
\nabla_{\beta_{kk'}^j} \mathrm{log\ q}(\theta_{kk'}^j | \alpha_{kk'}^j, \beta_{kk'}^j) = \\ \mathrm{log}\ (1-\theta_{kk'}^j)  + \Psi( \alpha_{kk'}^j + \beta_{kk'}^j) - \Psi( \beta_{kk'}^j )
\end{multline}
\begin{equation}
\nabla_{p_{ik}} \mathrm{log\ q}(z_i | \mathbf{p}_{i})  = \frac{z_{ik}}{p_{ik}}
\end{equation}
\begin{equation}
\nabla_{d_k} \mathrm{log\ q}(\pi | \mathbf{d})  = \mathrm{log}\ d_k - \Psi(d_k) -  \Psi(\sum_{k=1}^{\mathcal{K}}d_k)
\end{equation}

\subsection{Constrained Parameters} All the estimated parameters must be positive to remain in their distribution domain. In fact, each vector $\mathbf{p}_i$ and the vector $\mathbf{d}$ must sum one. We used the soft-plus function and a normalized soft-plus function to deal with these constraints.

\end{document}